\documentclass[11pt]{article}

\usepackage[final]{acl}

\usepackage{times}
\usepackage{latexsym}

\usepackage[T1]{fontenc}

\usepackage[utf8]{inputenc}
\usepackage{booktabs}
\usepackage{subcaption}   
\usepackage{caption}
\usepackage{algorithm}
\usepackage{algpseudocode}
\usepackage{amsmath}
\usepackage{amssymb}
\usepackage{wrapfig}
\usepackage{array}
\usepackage{tabularx}
\usepackage{hyperref}
\usepackage{url}
\usepackage[export]{adjustbox} 

\usepackage{microtype}

\usepackage{inconsolata}

\usepackage{graphicx}

\title{TrigReason: Trigger-Based Collaboration between Small and Large Reasoning Models}

\author{Yi Zhao$^{1,2}$, Yajuan Peng$^{3}$, Cam-Tu Nguyen$^{4}$, \\ \textbf{Zuchao Li$^{5,*}$, Xiaoliang Wang$^{4}$, Xiaoming Fu$^{6,3}$, Hai Zhao\textsuperscript{1,2,}} \thanks{$\ $  Corresponding author. This research was supported by the Shanghai Jiao Tong University 2030 Initiative, the Major Program of Chinese National Foundation of Social Sciences under Grant ‘The Challenge and Governance of Smart Media on News Authenticity’ [No. 23\&ZD213], the National Natural Science Foundation of China (Grant No. W2531047) and the National Natural Science Foundation of China (No. 62306216).}\\
\textsuperscript{1}AGI Institute, School of Computer Science, Shanghai Jiao Tong University, Shanghai, China\\
\textsuperscript{2}Key Laboratory of Shanghai Education Commission for Intelligent Interaction and \\ Cognitive Engineering, Shanghai Jiao Tong University, Shanghai, China \\
\textsuperscript{3}Shanghai Key Laboratory for Intelligent Information Processing, Fudan University\\
\textsuperscript{4}State Key Laboratory for Novel Software Technology, Nanjing University\\
\textsuperscript{5}School of Artificial Intelligence, Wuhan University\\
\textsuperscript{6}Institute of Computer Science, University of Göttingen, Göttingen, Germany
}

\begin{document}
\maketitle
\begin{abstract}

Large Reasoning Models (LRMs) achieve strong performance on complex tasks through extended chains of thought but suffer from high inference latency due to autoregressive reasoning. Recent work explores using Small Reasoning Models (SRMs) to accelerate LRM inference.
In this paper, we systematically characterize the capability boundaries of SRMs and identify three common types of reasoning risks: (1) path divergence, where SRMs lack the strategic ability to construct an initial plan, causing reasoning to deviate from the most probable path; (2) cognitive overload, where SRMs fail to solve particularly difficult steps; and (3) recovery inability, where SRMs lack robust self-reflection and error correction mechanisms. To address these challenges, we propose TrigReason, a trigger-based collaborative reasoning framework that replaces continuous polling with selective intervention. TrigReason delegates most reasoning to the SRM and activates LRM intervention only when necessary—during initial strategic planning (strategic priming trigger), upon detecting extraordinary overconfidence (cognitive offload trigger), or when reasoning falls into unproductive loops (intervention request trigger). The evaluation results on AIME24, AIME25, and GPQA-D indicate that TrigReason matches the accuracy of full LRMs and SpecReason, while offloading 1.70×–4.79× more reasoning steps to SRMs. Under edge–cloud conditions, TrigReason reduces latency by 43.9\% and API cost by 73.3\%. Our code is available at \href{https://github.com/QQQ-yi/TrigReason}{https://github.com/QQQ-yi/TrigReason}
\end{abstract}

\section{Introduction}

Large Reasoning Models (LRMs)~\citep{chatgpt-o1,deepseekai2025deepseekr1incentivizingreasoningcapability} have recently emerged as a powerful paradigm for tackling complex problem by leveraging extended chains of thought (CoT)~\citep{wei2022chain,yao2024tree,yao-etal-2024-got} during inference. Unlike standard large language models (LLMs) that directly generate output tokens, LRMs performs an internal reasoning process by generating a sequence of thinking tokens, which break down the input question into intermediate reasoning steps prior to producing the final answer. This structured reasoning behavior enables state-of-the-art performance across diverse domains such as mathematical reasoning~\citep{QwQ-32B}, code generation~\citep{ahmad2025opencodereasoningadvancingdatadistillation}, and agent~\citep{kimiteam2025kimik2openagentic}. However, this enhanced reasoning capacity comes at a significant cost: the autoregressive generation of long CoT sequences, often spanning thousands of thinking tokens, leads to prolonged response delays. This limitation has driven recent research into accelerating LRM inference.

Previous approaches to reasoning efficiency have primarily focused on refining the effective density of CoT to mitigate redundant or excessive reasoning~\citep{zhao2025smallkv}. Among these, reinforcement learning with a length penalty is widely adopted to encourage concise and effective reasoning trajectories~\citep{luo2025o1prunerlengthharmonizingfinetuningo1like,yang2025thinkneedselfadaptivechainofthought}. Alternative methods explore supervised fine-tuning using variable-length CoT data to promote efficient inference~\citep{xia2025tokenskipcontrollablechainofthoughtcompression,kang2024c3otgeneratingshorterchainofthought,ma2025cotvalvelengthcompressiblechainofthoughttuning}. Moreover, prompt engineering also have been proposed to guide models toward more streamlined reasoning through carefully designed input prompts~\citep{wu2025concisereasoningbiggains,xu2025chaindraftthinkingfaster,zhao2025dac}. Although these approaches enhance inference efficiency, they typically impose a reduced token budget for reasoning, which may lead to skipping critical logical steps or preventing necessary self-correction in the reasoning process. A separate strand of work aims to develop small language models with strong reasoning capabilities~\citep{dang2025reinforcementlearningreasoningsmall,zhao2025iam}. However, these methods may also suffer performance degradation due to the limited capabilities of small models.

\begin{figure*}[!htb]
  \centering
  \begin{subfigure}[t]{0.425\textwidth}
    \centering
    \includegraphics[scale=0.354,valign=t]{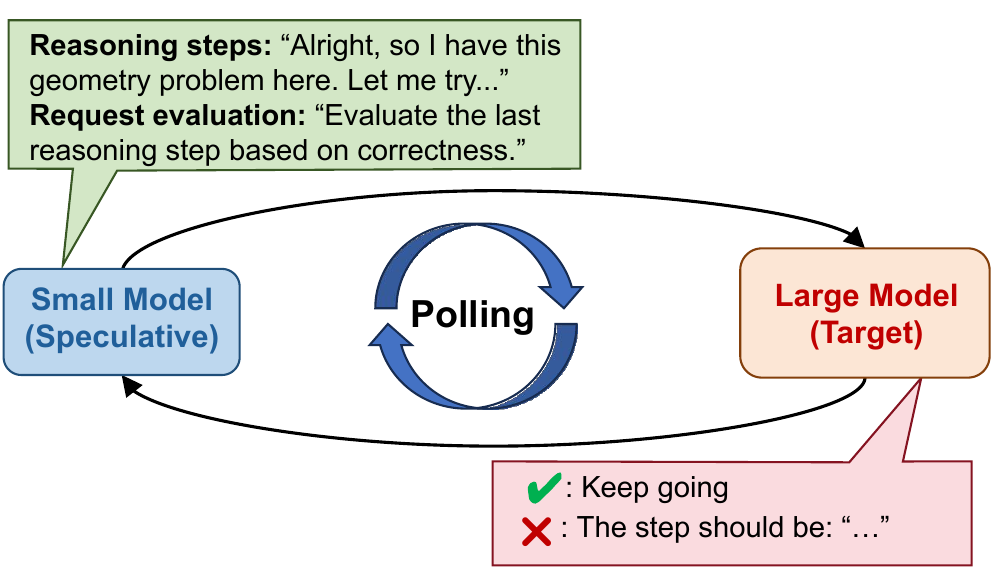}
    \caption{SpecReason framework.}
    \label{fig:spec_ill}
  \end{subfigure}
  \hfill
  \begin{subfigure}[t]{0.565\textwidth}
    \centering
    \includegraphics[scale=0.354,valign=t]{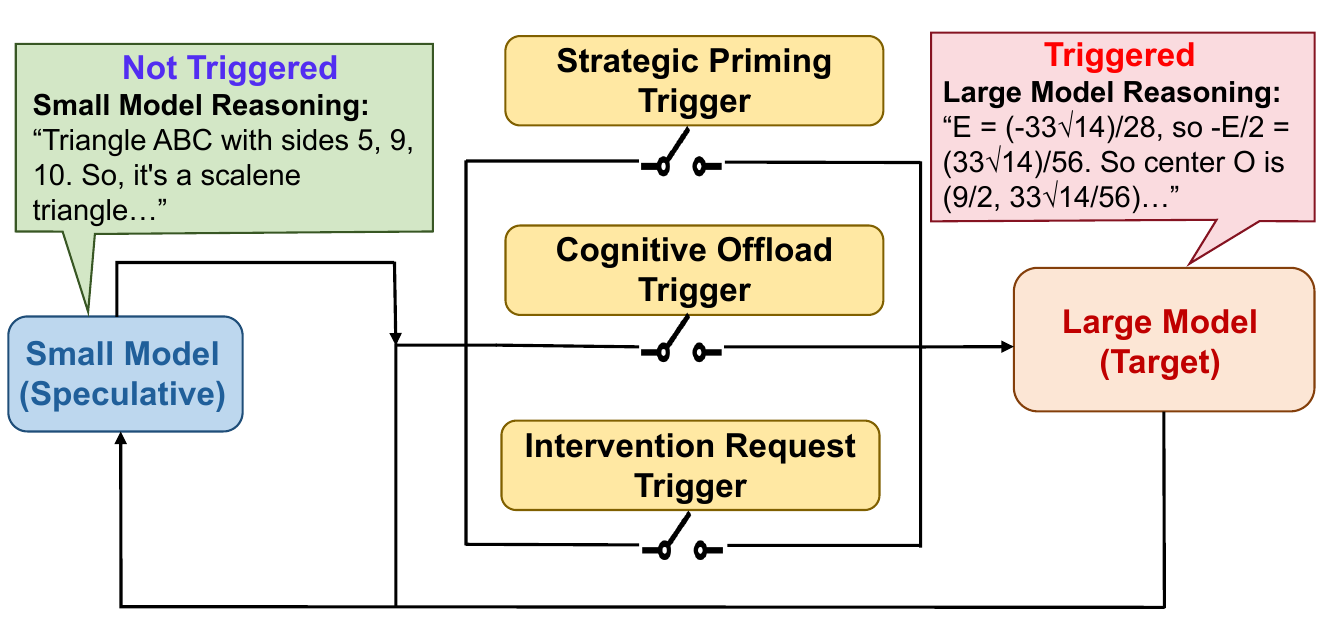}
    \caption{TrigReason framework.}
    \label{fig:trigger_ill}
  \end{subfigure}
  \vspace{1em} 
  \begin{minipage}{0.8\textwidth}
    \centering
    \includegraphics[scale=0.3,valign=t]{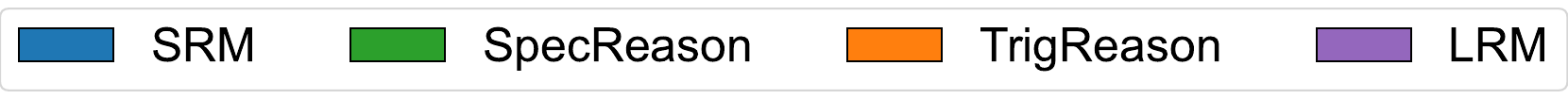} 
  \end{minipage}\vspace{-8pt}
  \begin{subfigure}[t]{0.29\textwidth}
    \centering
    \includegraphics[scale=0.29,valign=t]{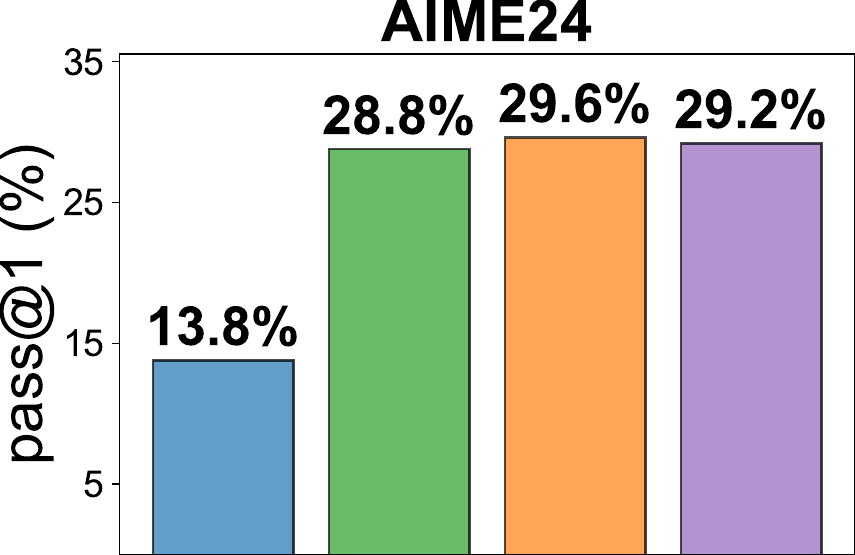}
    \caption{Accuracy comparison.}
    \label{fig:inst_acc}
  \end{subfigure}
  \hspace{1pt}
  \begin{subfigure}[t]{0.37\textwidth}
    \centering
    \includegraphics[scale=0.29,valign=t]{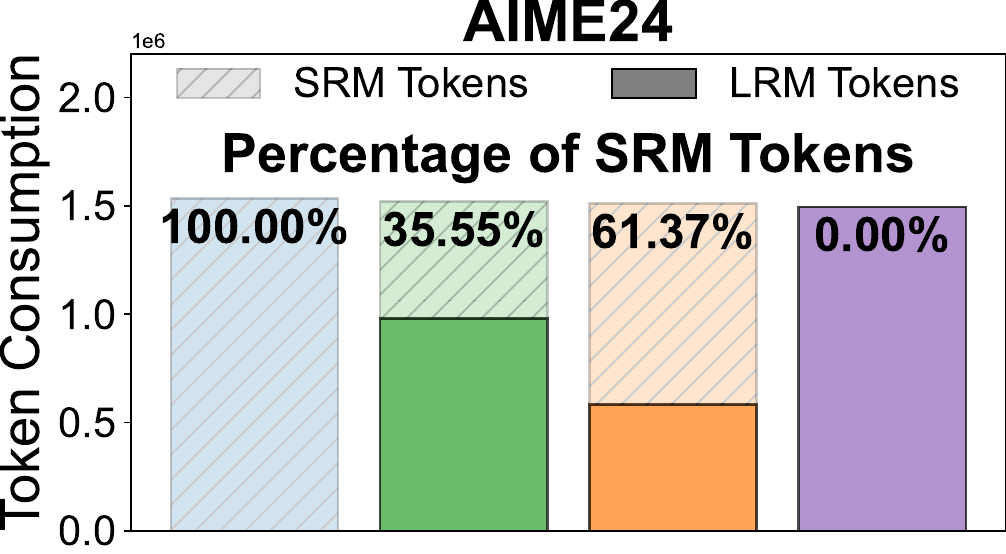}
    \caption{Token usage percentage.}
    \label{fig:inst_perc}
  \end{subfigure}
  \hspace{-1pt}
  \begin{subfigure}[t]{0.29\textwidth}
    \centering
    \includegraphics[scale=0.29,valign=t]{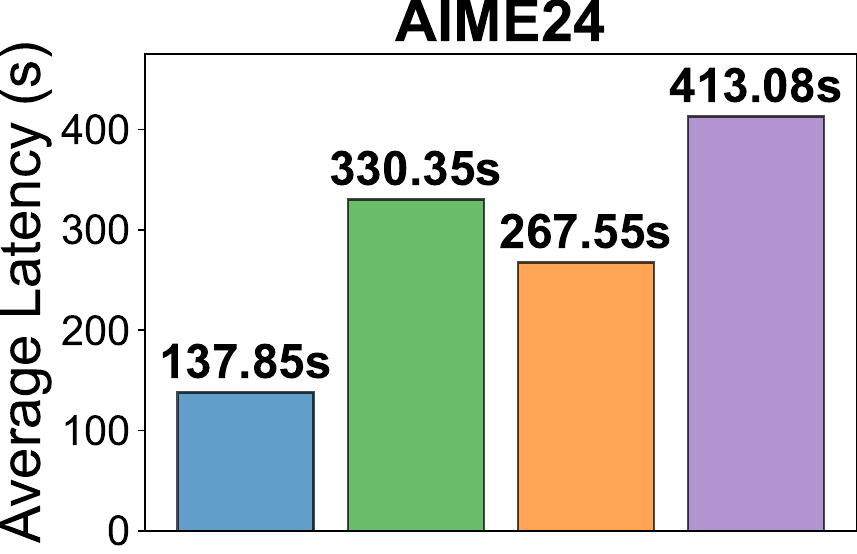}
    \caption{Latency comparison.}
    \label{fig:inst_latency}
  \end{subfigure}
  \caption{Overview of the reasoning frameworks and performance evaluation between SpecReason and TrigReason. The evaluation is on AIME24 benchmark using DeepSeek-R1-1.5B as SRM and QwQ-32B as LRM}
  \vspace{-15pt}
  \label{fig:overview}
\end{figure*}

Recently, SpecReason~\citep{pan2025specreasonfastaccurateinferencetime} observes that many LRM reasoning steps can be semantically covered by small reasoning model (SRM), and proposes a speculative paradigm that verifies SRM-generated steps via an LRM-as-a-Judge mechanism. While SpecReason can effectively reduce latency, it faces two key limitations. First, the LRM’s judgment is often unreliable(\S\ref{sec:unreliability}). Model judgment is inherently subjective due to behavior biases ingrained during training. Besides, evaluating individual reasoning steps is challenging, as the chain of thought remains incomplete. Second, as illustrated in Figure~\ref{fig:spec_ill}, its polling-based design, where the LRM is invoked at every reasoning step to validate the SRM’s output regardless of the complexity, resulting in significant overhead, especially in edge-cloud collaboration (\S\ref{sec:inefficiency}). These limitations lead to inefficient speculative reasoning, as excessive LRM intervention results in the final output being dominated by LRM-generated corrections rather than SRM reasoning.



These inefficiencies originate from an incomplete understanding of when and why SRM fails. Existing methods resort to frequent and blind verification, sacrificing efficiency for effectiveness. In this paper, we first characterize the capability boundaries of the SRM to identify the most common reasoning errors: path divergence risk, cognitive overload risk, and recovery inability risk (\S\ref{sec:risky}). Based on this analysis, we propose \textbf{TrigReason}, a event-triggered collaborative reasoning framework that shifts LRM correction from polling to selective intervention. As shown in Figure~\ref{fig:trigger_ill}, instead of continuous verification, TrigReason allows the SRM to reason autonomously until one of three purpose-designed triggers fires: (1) a strategic priming step from the LRM at the start, (2) a cognitive offload trigger when confidence becomes extraordinary, or (3) an intervention request when the SRM detects stagnant reasoning loops. As shown in Figure~\ref{fig:inst_acc}, \ref{fig:inst_perc} and \ref{fig:inst_latency}, this shift enables TrigReason to significantly increase the proportion of tokens generated by SRM (from 35.55\% to 61.37\% of total token consumption) while maintaining accuracy and substantially reducing end-to-end latency.


We evaluate TrigReason extensively on three challenging reasoning benchmarks, AIME24~\citep{aime2024dataset}, AIME25~\citep{aime2025dataset}, and GPQA Diamond~\citep{rein2024gpqa} across diverse SRM-LRM combinations. Results show that TrigReason maintains accuracy compared with both the full LRM and SpecReason, while utilizing 1.70$\times$ to 4.79$\times$ more SRM-generated tokens than SpecReason, indicating significantly higher reasoning steps offloading efficiency. Under edge-cloud collaboration scenarios, TrigReason achieves reduction of 43.9\% in latency and 73.3\% in API cost compared to SpecReason. These results demonstrate that TrigReason establishes a more effective paradigm for collaborative reasoning between small and large models, achieving significant improvements in inference efficiency without compromising accuracy.

\section{Motivation}

Recent advances in speculative reasoning have shown that collaboration between SRM and LRM can accelerate inference without sacrificing solution quality. SpecReason~\citep{pan2025specreasonfastaccurateinferencetime} exemplifies this progress by adopting an LRM-as-a-Judge framework, where the LRM is prompted to score each reasoning step generated by the SRM, determining whether to accept or reject it. In this section, we explore two key limitations hindering its practical effectiveness:

\textbf{Unreliable LRM judgment}: (1) the inherent biases of each model, which make the judge prone to subjectivity rather than serving as an objective detector of errors; and (2) the difficulty of assessing intermediate reasoning steps when chains of thought are not yet fully formed, thus often unclear.

\textbf{Inefficiency of polling-based execution}: the step-level granularity of LRM invocation leads to frequent communication and computation overhead, especially in edge-cloud collaboration.

These limitations undermine intended acceleration benefits of speculative reasoning, as the majority of the final output is derived from the corrections of LRM.

\subsection{Unreliability of LRM Judgment}
\label{sec:unreliability}

While SpecReason advances speculative inference efficiency in reasoning acceleration, its LRM-as-a-Judge mechanism suffers from a critical flaw, as LRM fails to reliably validate the correctness of reasoning steps from SRM. 

\begin{figure*}[!htb]
  \centering
  \begin{subfigure}[t]{0.23\textwidth}
    \centering
    \includegraphics[scale=0.23]{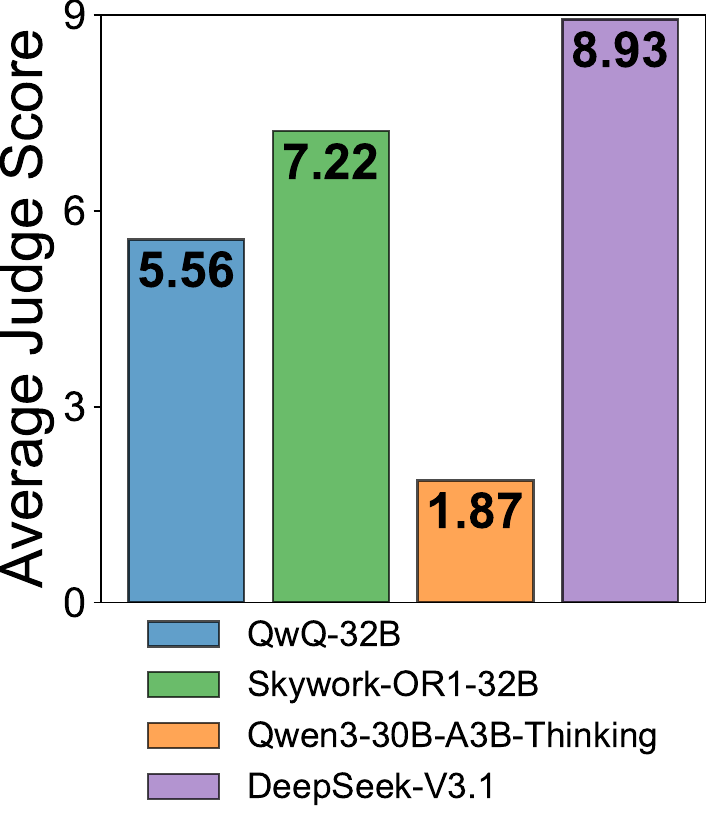}
    \caption{}
    \label{fig:non-consis}
  \end{subfigure}
    \begin{subfigure}[t]{0.38\textwidth}
    \centering
    \includegraphics[scale=0.22]{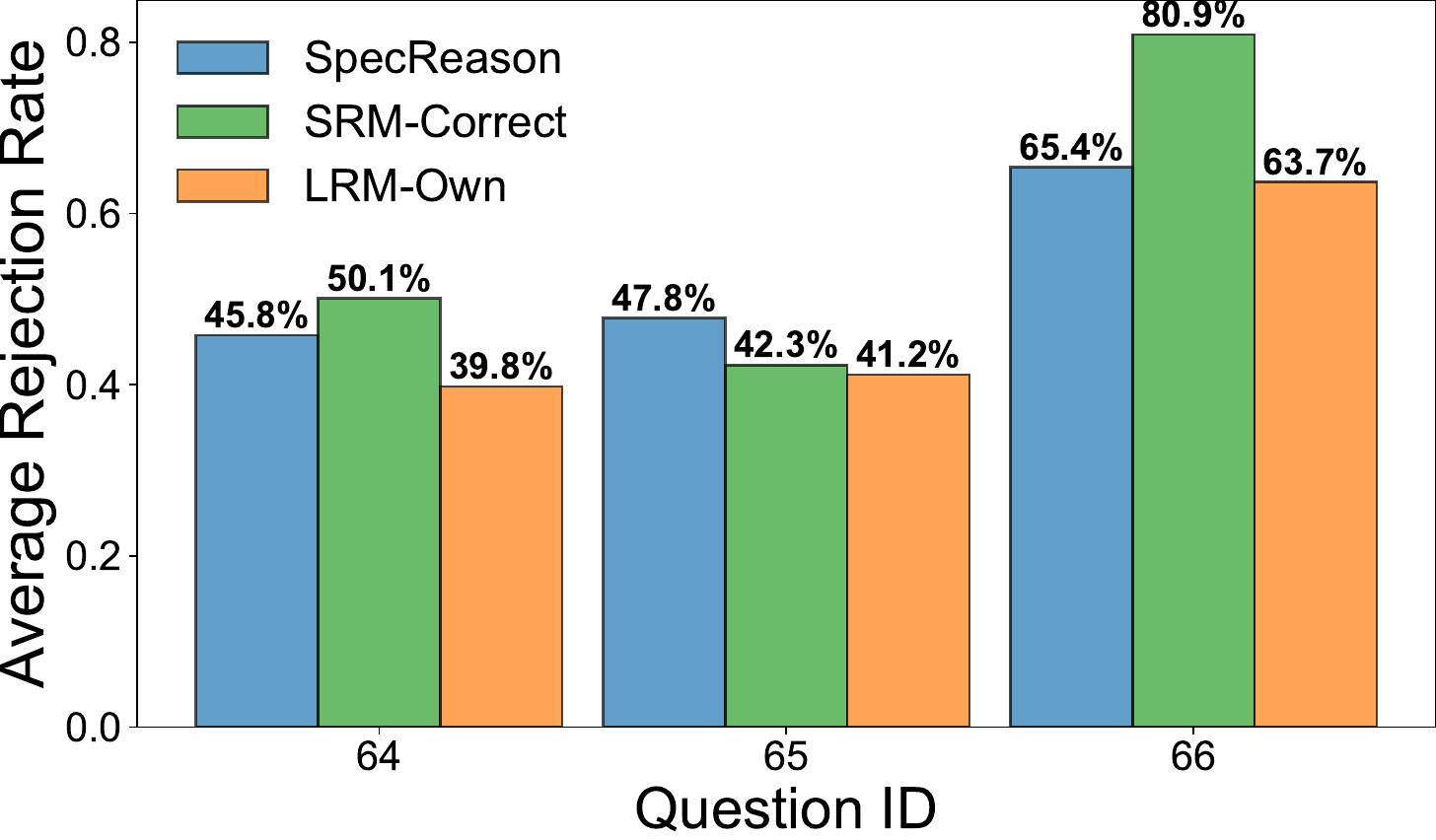}
    \caption{}
    \label{fig:disting}
  \end{subfigure}
  \begin{subfigure}[t]{0.18\textwidth}
    \centering
    \includegraphics[scale=0.24]{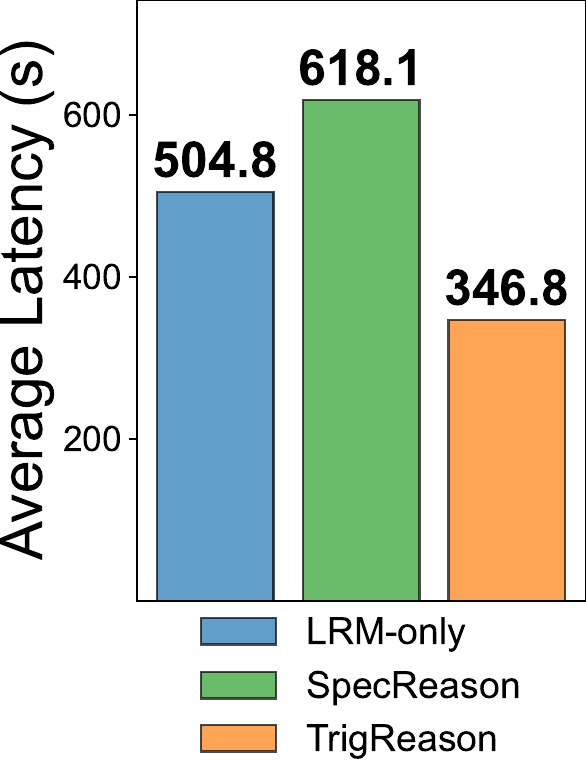}
    \caption{}
    \label{fig:overhead1}
  \end{subfigure}
    \begin{subfigure}[t]{0.18\textwidth}
    \centering
    \includegraphics[scale=0.24]{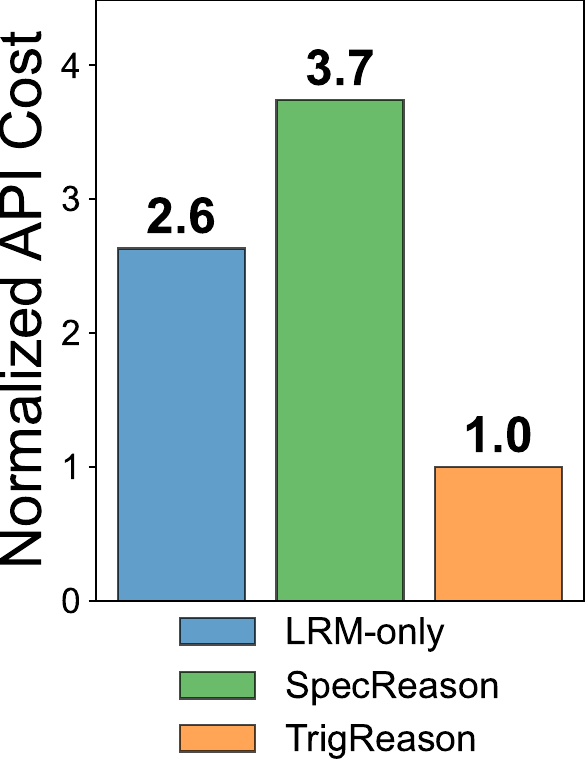}
    \caption{}
    \label{fig:overhead2}
  \end{subfigure}
  \caption{(a) Average judge scores of same trajectories from four different LRM, showing extreme inter-judge inconsistency. (b) Average rejection rate on three reasoning trajectories. Even for correct steps and LRM-own generation, LRM judgment still shows high-level rejection rate. (c) and (d) are comparison of average latency and API cost in edge-cloud collaborative reasoning, respectively.}
  \label{fig:overview}
\end{figure*}

Due to inherent biases in model training, LRM judgments are often preference-driven, leading to subjective judgments for the same content across different models. We evaluate identical reasoning trajectories using four different LRMs. As shown in Figure \ref{fig:non-consis}, the LRMs assign widely divergent scores to the same trajectory (from 1.87 to 8.93). This polarization indicates that LRM judgments are heavily influenced by model-specific priors, rather than objective reasoning quality.

Furthermore, to assess the difficulty of verifying intermediate reasoning steps in incomplete chains of thought, we sample reasoning trajectories from the AIME24 dataset. We categorize these trajectories into three types (defined below), and evaluate SpecReason to quantify the unreliability verification, using QwQ-32B as the LRM and DeepSeek-R1-1.5B as the SRM:
\begin{itemize}
\item \textbf{SpecReason}: trajectories of SpecReason with mixed steps from SRM and LRM;
\item \textbf{SRM-Correct}: trajectories from the SRM that yield correct final answers;
\item \textbf{LRM-Own}: trajectories generated independently by the LRM itself.
\end{itemize}
We follow SpecReason’s experimental setup: the LRM assigns scores in the range [0, 9], with a threshold of 7 for acceptance, and each question is evaluated over 16 runs to compute the average rejection rate across reasoning trajectories. Figure \ref{fig:disting} presents results on three questions where the SRM can correctly solve, as the remaining results are shown in Appendix \ref{app:judge}. The results reveal that the LRM rejects \textbf{50.1\% to 80.9\%} of correct and valid reasoning steps generated by the SRM. More surprisingly, the LRM even rejects up to \textbf{63.7\%} of its own generation.

These results indicate that the LRM-as-a-Judge paradigm is unreliable in verifying reasoning steps. Due to discrepancies of model inherent biases and the difficulty of assessing intermediate reasoning steps, the LRM is prone to regenerate the correct draft of SRM that differ only in phrasing or reasoning path. This unreliable judgment forces excessive LRM intervention to ensure solution quality, significantly undermining the efficiency of speculative reasoning.

\vspace{-8pt}

\subsection{Inefficiency of Polling-based Execution}
\label{sec:inefficiency}

SpecReason adopts polling-based execution that requires the LRM to intervene at every reasoning step, ignoring both step complexity and the SRM’s internal confidence. Frequent LRM calls incur substantial overhead, and also diminish the expected efficiency gains of speculative reasoning.

In edge-cloud setups, a representative deployment for speculative reasoning, the SRM executes on resource-constrained edge devices while the LRM operates in the cloud. Frequent polling under this architecture induces significant \textbf{network round-trips and API costs}, exacerbating system-level inefficiencies. We implement a edge-cloud deployment to quantify the effect, as the SRM (DeepSeek-R1-1.5B) runs locally and the LRM is accessed via DeepSeek API. As shown in Figure \ref{fig:overhead1} and \ref{fig:overhead2}, even compared to LRM-only execution, SpecReason exhibits lower efficiency, with a 22.44\% increase in latency and a 42.31\% higher API cost.

\section{method}

To address the shortcomings of polling-based LRM verification, we propose TrigReason, guided by two key ideas: (1) studying how SRM fails in order to identify the most common reasoning risks, thereby enabling more objective targeting and reducing blind reliance on LRM judges; and (2) triggering LRM for speculative reasoning correction based on event signals rather than at every step, which reduces the number of LRM calls and significantly lowers latency. Allowing SRM to generate the reasoning chain to some extent before intervention also makes the reasoning path more explicit, enabling LRM to deliver more effective corrections.


\subsection{Characterization of SRM Reasoning Risks}
\label{sec:risky}

The limitations of current speculative reasoning originate from an insufficient understanding of when and why SRM fails, resulting in an inability to distinguish between harmless reasoning variations and high-risk steps. By characterizing the capability boundaries of the SRM, LRM intervention can be reserved only for critical steps, avoiding excessive verification and missed interventions. To address this issue, we conduct a systematic analysis of reasoning trajectories generated by SRM and identify three core failure modes that cause distinct capability gap between SRM and LRM: path divergence risk, cognitive overload risk, and recovery inability risk.

\begin{figure*}[!htb]
\centering
\includegraphics[scale=0.53]{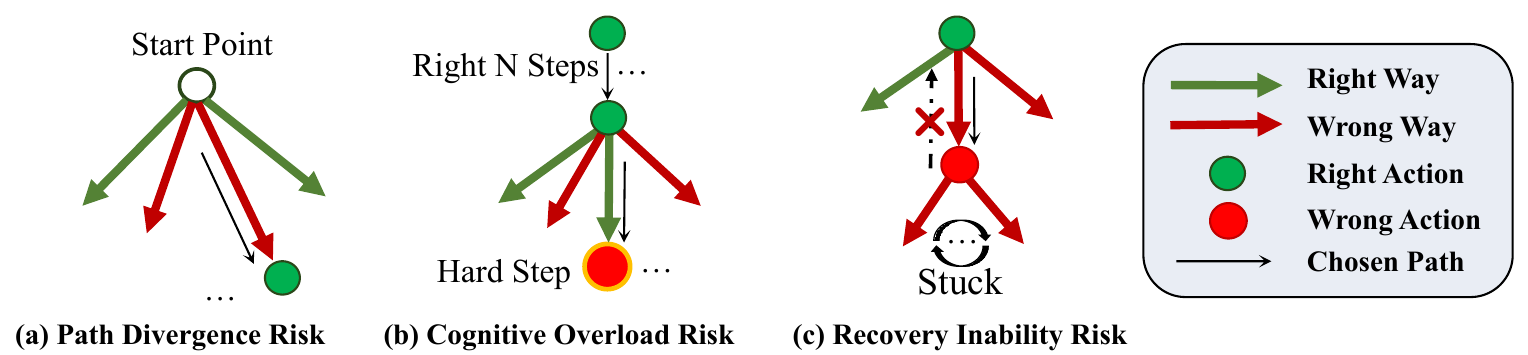}
\caption{Illustration of three typical risk patterns reflecting the SRM-LRM capability gap.}
\vspace{-15pt}
\label{fig:risk}
\end{figure*}

To identify the critical steps and risk patterns, we compared correct and incorrect reasoning trajectories through different scaled reasoning models on the AIME24 datasets. Figure \ref{fig:risk} shows three typical risk patterns, which characterize the capability gap between different model scales. Examples of the three risks are shown in Appendix \ref{app:case_risk}.

\textbf{(1) Path Divergence Risk: }occurs at the beginning of reasoning process, representing a fault-oriented solution branch. Unlike factual hallucinations or arithmetic mistakes, it arises from the SRM’s failure to decompose the problem or anticipate the implications of alternative approaches.

As the study case shown in Appendix \ref{app:case_risk}, SRMs often jump to computation or apply familiar but unsuitable methods, whereas larger models first analyze problem structure and plan strategically. The observation highlights the lack of strategic foresight in smaller models during initial planning.

\textbf{Insight:} let the LRM generate initial reasoning steps for strategic planning or problem decomposition, enabling the SRM to efficiently execute downstream steps under a validated reasoning path.

\textbf{(2) Cognitive Overload Risk: }occurs in a subset of specific steps that demand high cognitive load, such as complex arithmetic computations requiring the retention of numerous intermediate states (e.g., multi-step fraction simplification, symbolic manipulation, or multi-hop logical inference). 

As the study case shown in Appendix \ref{app:case_risk}, although relatively infrequent, these errors are highly consequential, leading cascading failures in subsequent reasoning steps. Yet for routine reasoning (e.g., interpretation, sequencing, and simple calculation), smaller models are reliable. 
Therefore, the key lies in identifying a insightful signal to detect cognitive overload in the SRM.

We analyze 160 reasoning trajectories from SRMs on 10 problems with significant SRM–LRM performance gaps. Among the 93 trajectories containing clear SRM incorrect reasoning steps, 94.6\% steps exhibit overconfident steps (i.e., over 85\% of tokens have perplexity $< 1.05$), compared to only 38.1\% overconfidence steps across all steps (see Appendix \ref{app:overconfidence} for details). This pattern shows SRM failure is often preceded by abnormally low token-level perplexity, indicating overconfident and deterministic generation. The overconfidence is not a sign of capability, but rather a symptom of mechanical pattern completion under cognitive overload.

\textbf{Insight:} SRMs are limited not by reasoning ability throughout the process, but by cognitive capacity. This pattern motivates a light-touch intervention strategy: leveraging overconfidence as a signal of cognitive overload to trigger LRM assistance, keeping the SRM on a correct reasoning trajectory.

\textbf{(3) Recovery Inability Risk: }occurs when minor errors or ambiguous interpretations lead the SRM to deviate from the correct path, resulting in increasingly incoherent reasoning. In contrast, LRM can implicitly reflect, detect anomalies, and backtrack to correct its approach. 

As the study case shown in Appendix \ref{app:case_risk}, SRMs lack the ability to detect deviations or initiate self-correction, causing them to persist on erroneous paths and eventually stagnate.

\textbf{Insight:} let LRM to reflect and re-guide the path-enabling corrective when signs of stagnation or contradiction are detected in the SRM’s reasoning trajectory.

\subsection{Event-triggered LRM intervention}

Based on the observations in \S\ref{sec:risky}, we propose TrigReason, a trigger-based framework for collaborative SRM-LRM execution. TrigReason introduces three targeted triggers to address critical failure risks in SRM reasoning: \textbf{strategic priming trigger}, \textbf{cognitive offload trigger}, and \textbf{intervention request trigger}. Owing to sparse yet crucial LRM intervention, TrigReason ensures high answer quality while enabling efficient and low-cost collaborative reasoning, as shown in Figure \ref{fig:overhead1} and \ref{fig:overhead2}.

\subsubsection{Strategic Priming Trigger}

The Strategic Priming Trigger is designed to address the Path Divergence Risk. By decoupling strategic planning from step-by-step execution, TrigReason uses the LRM to perform initial reasoning, ensuring the SRM begins on a valid and coherent trajectory.

Specifically, given an input question $x$, we first sample the first $n$ reasoning steps from the LRM $L$:
\begin{equation}
    y_{1:n} \sim p_L(y_{1:n} \mid x),
\end{equation}
where $p_L$ denotes the conditional distribution of the LRM, and $n$ is a pre-defined priming steps. 
After this priming phase, control is transferred to the SRM $S$, which continues the reasoning chain:
\begin{equation}
    y_t \sim p_S(y_t \mid y_{<t}, x), \quad \text{for } t > n.
\end{equation}

\subsubsection{Cognitive Offload Trigger}
The Cognitive Offload Trigger aims to address the Cognitive Overload Risk. TrigReason leverages the extraordinary overconfidence of SRM as an early warning signal to trigger LRM intervention at critical junctures. 

To quantify this behavior, we define the token-level perplexity at position $t$ as:
\begin{equation}
    \textit{PPL}(t) = \exp\left(-\log p_S(y_t \mid y_{<t}, x)\right),
\end{equation}
where $p_S(y_t \mid y_{<t}, x)$ is the probability assigned by the SRM to token $y_t$, given the prefix $y_{<t}$ and input $x$. For a given reasoning step $s$, let $T_s$ denote the set of token positions in $s$. We compute the low-perplexity ratio $r_s$ as the fraction of tokens in $s$ with perplexity below a threshold $\tau$:
\begin{equation}
    r_s = \frac{1}{|T_s|} \sum_{t \in T_s} \mathbf{1}\left[\textit{PPL}(t) < \tau\right],
\end{equation}
where $\tau$ is a sensitivity threshold and $\mathbf{1}\left[\cdot\right]$ is the indicator function, equal to 1 if the condition is true, and 0 otherwise. The Cognitive Offload Trigger fires when $r_s > \rho$, where $\rho$ is a coverage threshold:
\begin{equation}
    \text{Trig}_{\text{cognitive}}(s) = \mathbf{1}\left[r_s > \rho\right].
\end{equation}
Upon activation, the current step $s$ is regenerated by the LRM:
\begin{equation}
    y_s \sim p_L(\cdot \mid y_{<s}, x).
\end{equation}


\subsubsection{Intervention Request Trigger}
The Intervention Request Trigger aims to mitigate the Recovery Inability Risk. TrigReason monitors for linguistic markers of reasoning stagnation, and invokes the LRM to realign the reasoning path upon detection. Obesevation indicates that, SRM often generates distinctive hesitation patterns (e.g., "wait", "hmm", "alternatively"), which reflects an implicit recognition of difficult reasoning steps.


We define a finite set $\mathcal{H}$ of hesitation words for detection (Appendix \ref{app:hesitation} includes the complete list).
At each reasoning step $s$, we determine whether the generation contains at least one token from $\mathcal{H}$:
\begin{equation}
    h_s = \mathbf{1}\left[ \exists\, y_t \in y_s \text{ such that } y_t \in \mathcal{H} \right].
\end{equation}
The Intervention Request Trigger fires when hesitation is observed in $k$ consecutive steps:
\begin{equation}
    \text{Trig}_{\text{intervention}} = \mathbf{1}\left[ \sum_{i=0}^{k-1} h_{s-i} = k \right],
\end{equation}

Upon activation, the system transfers control to the LRM for the next $m$ steps:
\begin{equation}
    y_{s+1:s+m} \sim p_L(\cdot \mid y_{\leq s}, x),
\end{equation}
then LRM is able to assess the current state, identify inconsistencies, and recorrect reasoning path. After $m$ steps, control returns to the SRM. The main algorithm of TrigReason is shown in Appendix \ref{sec:algorithm}.


\section{Experiments}

\subsection{Setup}

\paragraph{Models.} \textbf{LRM}: QwQ-32B~\citep{QwQ-32B} (32B dense model) and Qwen3-30B-A3B-Thinking-2507~\citep{qwen3technicalreport} (30B MoE model, 3B active). \textbf{SRM}: DeepSeek-R1-1.5B~\citep{deepseekai2025deepseekr1incentivizingreasoningcapability} and Qwen3-0.6B~\citep{qwen3technicalreport}. Both models are equipped with CoT reasoning capabilities. We conduct experiments across four SRM-LRM pairings to evaluate the generalization of TrigReason under diverse model architectures and scales. In the edge-cloud deployment setting, the LRM (DeepSeek-V3.1, 671B MoE) is accessed via DeepSeek API~\citep{deepseek_api}.

\paragraph{Datasets.}\textbf{AIME24}~\citep{aime2024dataset} and \textbf{AIME25}~\citep{aime2025dataset} are high-school math competition problems requiring multi-step algebraic and combinatorial reasoning. \textbf{GPQA Diamond}~\citep{rein2024gpqa} is a graduate-level multiple-choice question set that covers advanced topics in physics, chemistry and biology, known for its high factual and logical complexity. We also evaluate TrigReason on additional reasoning domains such as logical and commonsense reasoning in Appendix \ref{app:generalization}.

\paragraph{Evaluation Metrics.} (1) \textbf{Accuracy}: following prior work~\citep{pan2025specreasonfastaccurateinferencetime}, we use \textbf{pass@1} with k = 16. Specifically, 16 responses are sampled for each question at temperature = 0.6, and the final accuracy is calculated as the average accuracy for every response. (2) \textbf{Efficiency}: as the total token consumption across methods is similar, we utilize the ratio of tokens generated by the SRM to the total reasoning tokens as a robust efficiency metric, donated as \textbf{SMT percentage}. We exclude latency from evaluation, as it is highly sensitive to hardware, system load, and scheduling variability, which could confound cross-method comparisons.

The method performance is visualized through the \textbf{Accuracy-Efficiency} plane, where the \textit{x} and \textit{y} axis represent \textit{SMT percentage} and \textit{pass@1}, respectively. Closer to the top-right performs better.

\paragraph{Baselines.} (1) \textbf{SpecReason}~\citep{pan2025specreasonfastaccurateinferencetime}, a polling-based collaborative method. (2) standalone reasoning framework using \textbf{only the SRM} or \textbf{only the LRM}.

\paragraph{Implementation Details.} All experiments are conducted on 8 NVIDIA RTX 4090 GPUs using SGLang v0.4.9~\citep{zheng2023efficiently} as the inference engine, with prefix caching and tensor parallelism (degree 4) enabled. We set the logprobs=True parameter during request to calculate token-level perplexity. Unless otherwise stated, generation uses temperature = 0.6 and top\_p = 0.95. The default token budget is 8192 tokens; for the impact of thinking budget analysis (Appendix \ref{sec:token_budget}), we evaluate budgets ranging from 2K to 32K. For TrigReason parameters, we set the priming step count $n = 20$ and rectification steps $m = 1$. The cognitive overload threshold $\rho$ is set to 0.85 for DeepSeek-R1-1.5B and 0.75 for Qwen3-0.6B with $\tau=0.85$. The rationale for these hyperparameter choices is justified through ablation studies in (\S\ref{sec:ablation}).

\begin{figure*}[!htb]
    \centering
    \begin{subfigure}[t]{0.7\textwidth}
        \centering
        \includegraphics[width=0.85\linewidth]{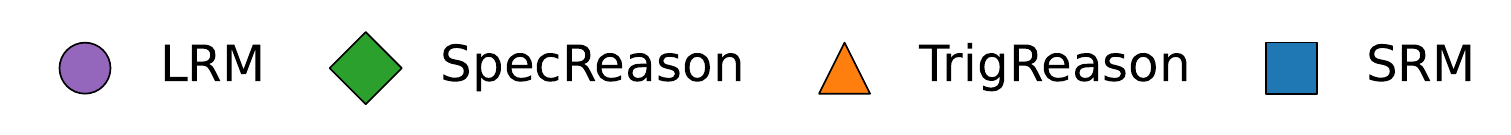}
    \end{subfigure}
    \begin{subfigure}[t]{0.32\textwidth}
        \centering
        \includegraphics[width=\linewidth]{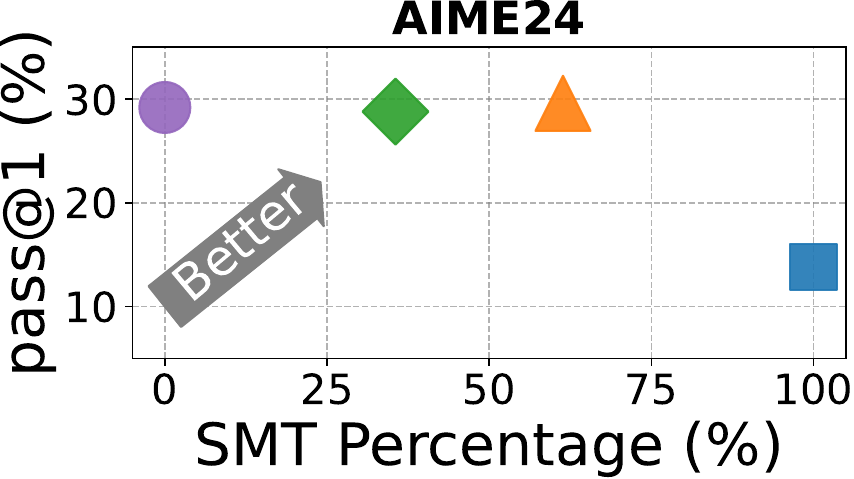}
    \end{subfigure}
    \hfill
    \begin{subfigure}[t]{0.32\textwidth}
        \centering
        \includegraphics[width=\linewidth]{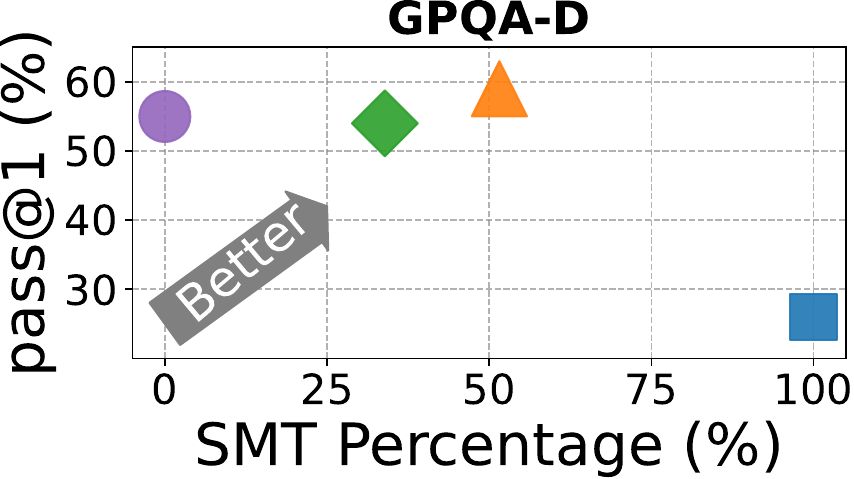}
    \end{subfigure}
    \hfill
    \begin{subfigure}[t]{0.32\textwidth}
        \centering
        \includegraphics[width=\linewidth]{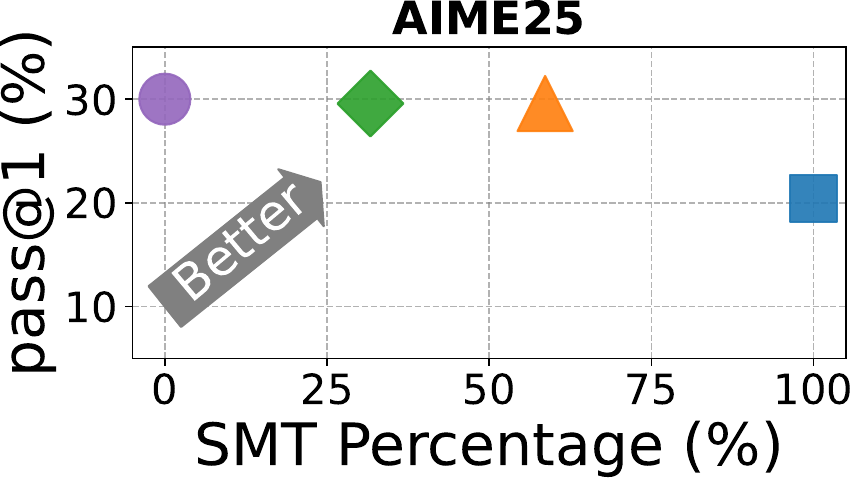}
    \end{subfigure}
    
    \caption*{(a) DeepSeek-R1-1.5B + QwQ-32B}
    
    \begin{subfigure}[t]{0.32\textwidth}
        \centering
        \includegraphics[width=\linewidth]{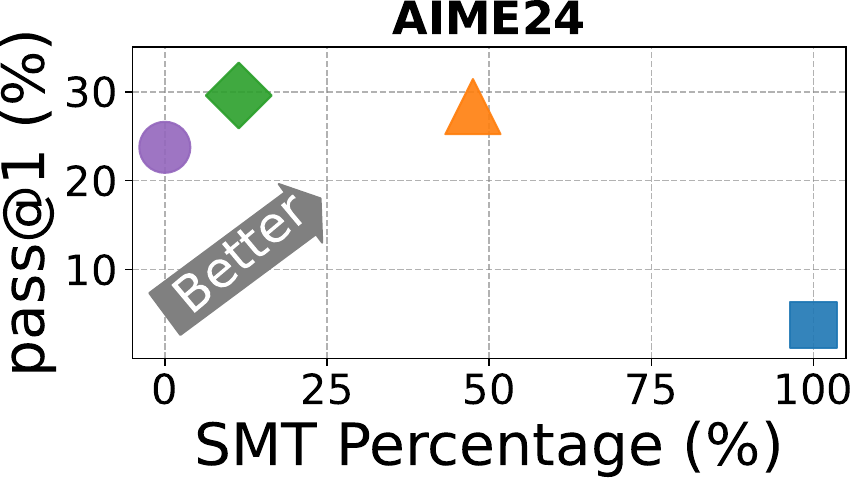}
    \end{subfigure}
    \hfill
    \begin{subfigure}[t]{0.32\textwidth}
        \centering
        \includegraphics[width=\linewidth]{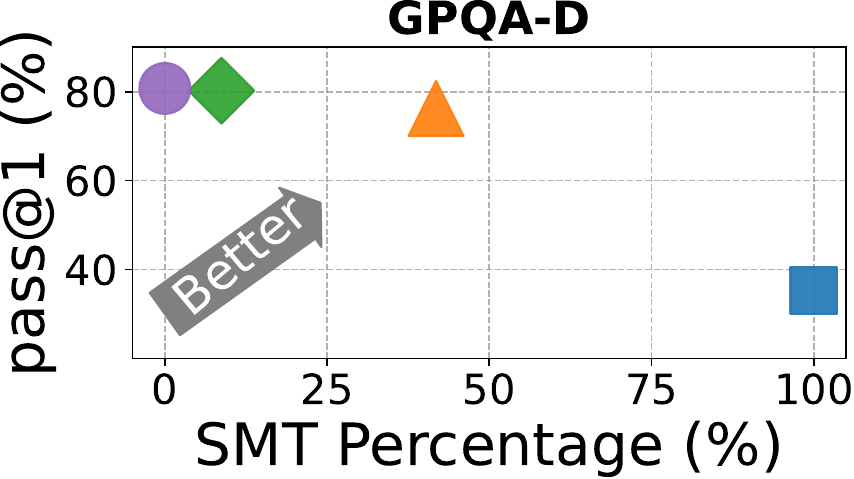}
    \end{subfigure}
    \hfill
    \begin{subfigure}[t]{0.32\textwidth}
        \centering
        \includegraphics[width=\linewidth]{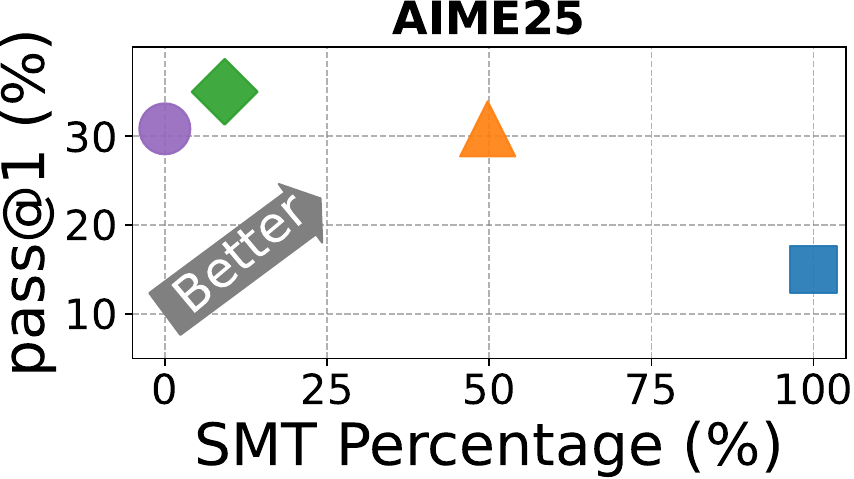}
    \end{subfigure}
    
    \caption*{(b) Qwen3-0.6B + Qwen3-30B-A3B-Thinking}
    
    \begin{subfigure}[t]{0.32\textwidth}
        \centering
        \includegraphics[width=\linewidth]{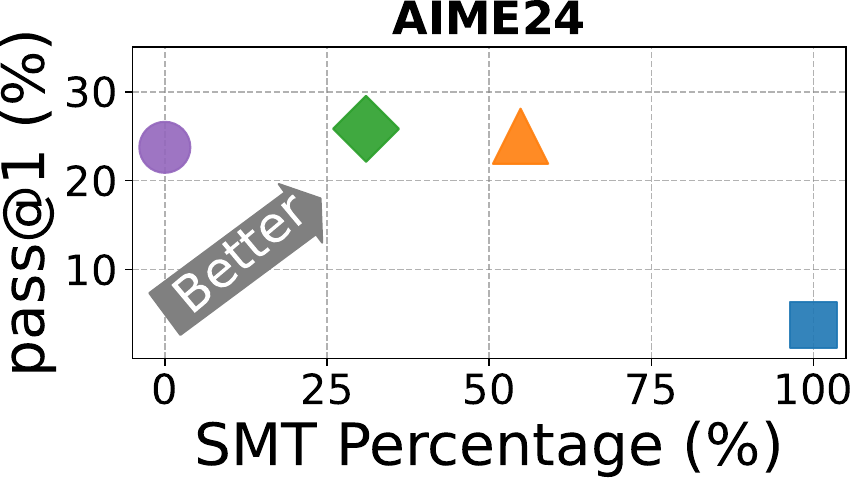}
    \end{subfigure}
    \hfill
    \begin{subfigure}[t]{0.32\textwidth}
        \centering
        \includegraphics[width=\linewidth]{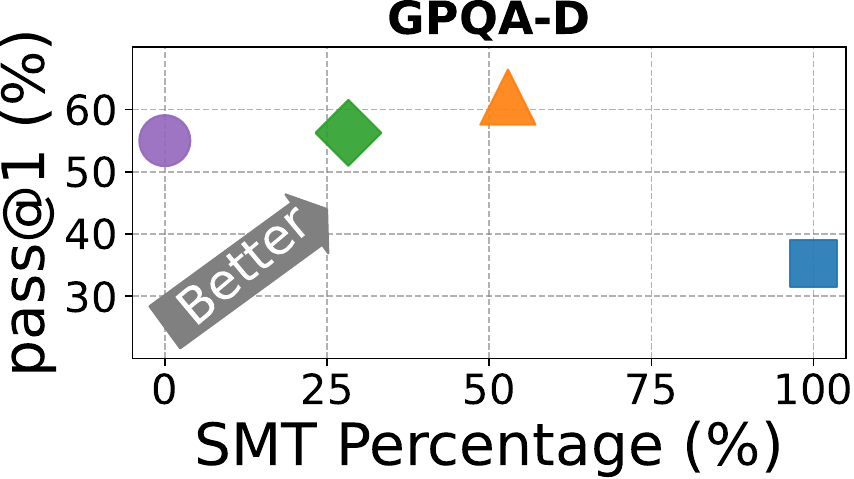}
    \end{subfigure}
    \hfill
    \begin{subfigure}[t]{0.32\textwidth}
        \centering
        \includegraphics[width=\linewidth]{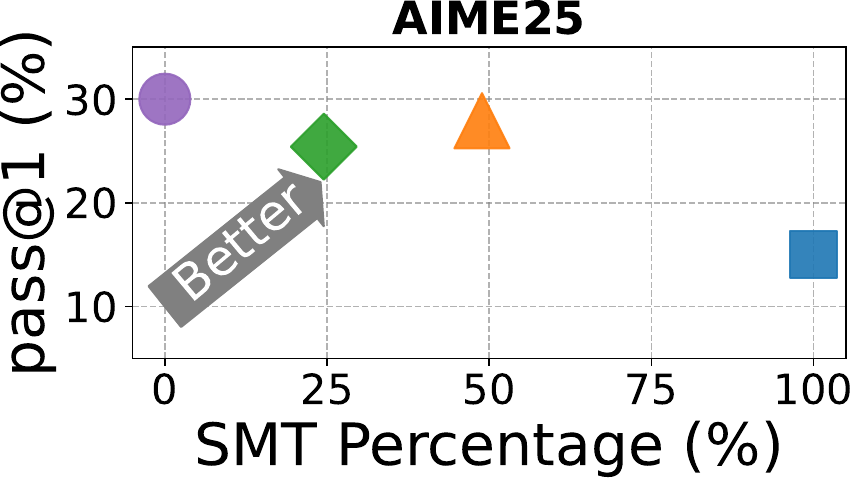}
    \end{subfigure}
    
    \caption*{(c) Qwen3-0.6B + QwQ-32B}
    
    \begin{subfigure}[t]{0.32\textwidth}
        \centering
        \includegraphics[width=\linewidth]{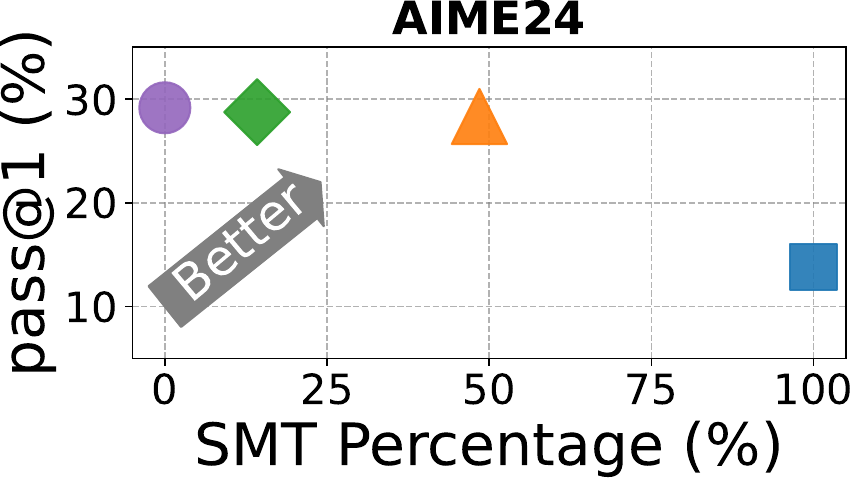}
    \end{subfigure}
    \hfill
    \begin{subfigure}[t]{0.32\textwidth}
        \centering
        \includegraphics[width=\linewidth]{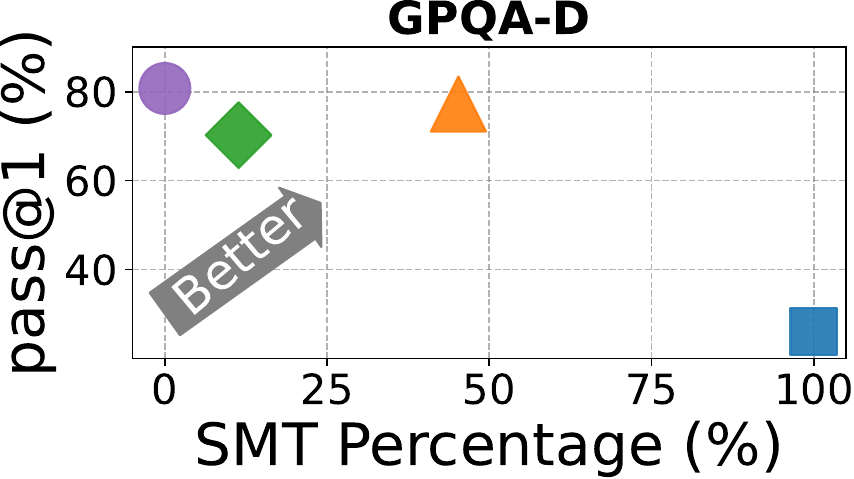}
    \end{subfigure}
    \hfill
    \begin{subfigure}[t]{0.32\textwidth}
        \centering
        \includegraphics[width=\linewidth]{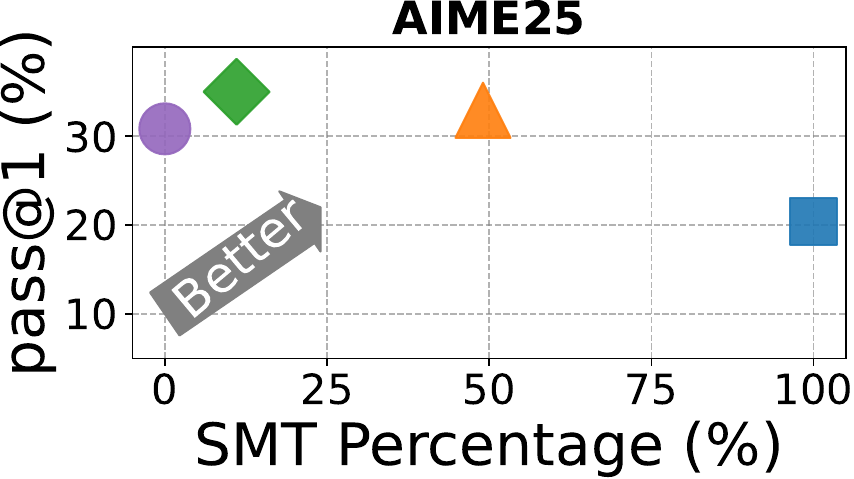}
    \end{subfigure}
    
    \caption*{(d) DeepSeek-R1-1.5B + Qwen3-30B-A3B-Thinking}
    \vspace{-5pt}
    \caption{Performance comparison across benchmarks and model combinations. The vertical axis shows accuracy (higher is better), and the horizontal axis shows the percentage of tokens generated by the SRM (SMT Percentage), reflecting reasoning efficiency (higher is more efficient). 
    }
    \label{fig:main_results}
\end{figure*}

\subsection{Main Results}

We evaluate TrigReason on AIME24, AIME25, and GPQA Diamond across four SRM-LRM combinations, comparing against vanilla LRM/SRM baselines and SpecReason. The results are shown in Figure~\ref{fig:main_results}.

\textbf{Stable Accuracy.} Despite offloading substantial reasoning to the SRM, TrigReason consistently matches or even exceeds LRM performance. On average, it achieves 105.8\% (AIME24), 104.7\% (AIME25), and 99.6\% (GPQA Diamond) of the LRM's accuracy across model pairs, with individual configurations (Qwen3-0.6B + Qwen3-30B-A3B-Thinking-
2507 on AIME24) reaching up to 119.3\%. In several cases, TrigReason surpasses the LRM baseline, suggesting that trigger-based intervention can yield robust and effective reasoning trajectories.

\textbf{Higher Efficiency.} While matching SpecReason in accuracy, TrigReason achieves significantly greater efficiency. On average, TrigReason utilizes $1.70\times-4.79\times$ more SRM tokens than SpecReason across benchmarks. Specifically, the SMT Percentage increases by $1.70\times$, $4.79\times$, $1.88\times$, and $3.94\times$ across the four combinations. This substantial gain indicates that TrigReason’s mechanism more effectively identifies and accepts valid reasoning steps.

\begin{figure}[!htb]
\centering
\includegraphics[scale=0.34]{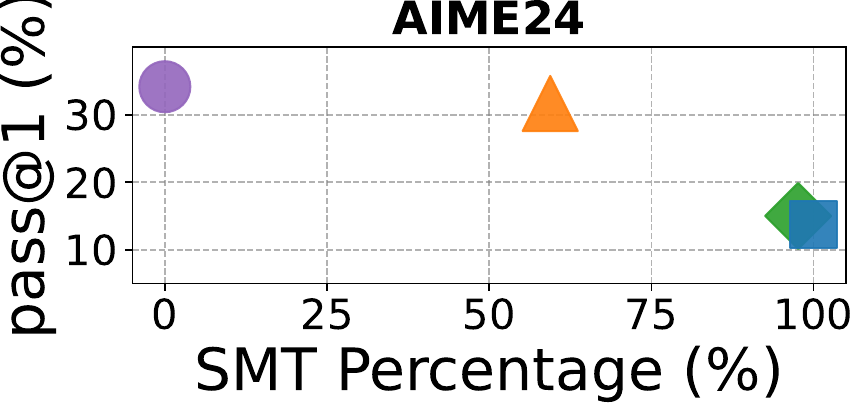}
\caption{Accuracy on AIME24 in an edge-cloud setup.}
\vspace{-15pt}
\label{fig:edge_co}
\end{figure}

In general, the results show that TrigReason achieves accuracy on par with LRM-only, while significantly improving efficiency through increased SRM utilization and less LRM calls. The evaluation results indicate that TrigReason achieves a superior efficiency-accuracy trade-off, representing a clear advancement in step-level speculative reasoning for collaborative inference.

\begin{figure*}[h]
    \centering
    \begin{subfigure}[t]{0.32\textwidth}
        \centering
        \includegraphics[scale=0.235]{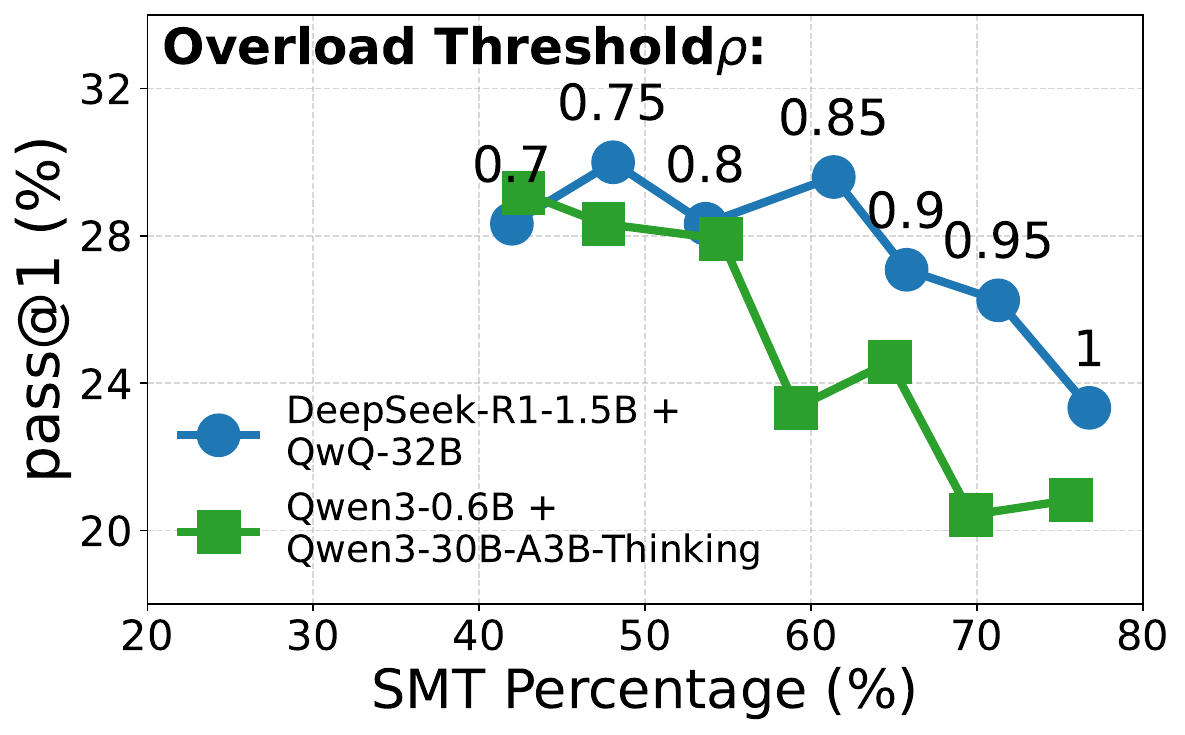}
        \caption{}
        \label{fig:ab1}
    \end{subfigure}
    \hfill
    \begin{subfigure}[t]{0.32\textwidth}
        \centering
        \includegraphics[scale=0.235]{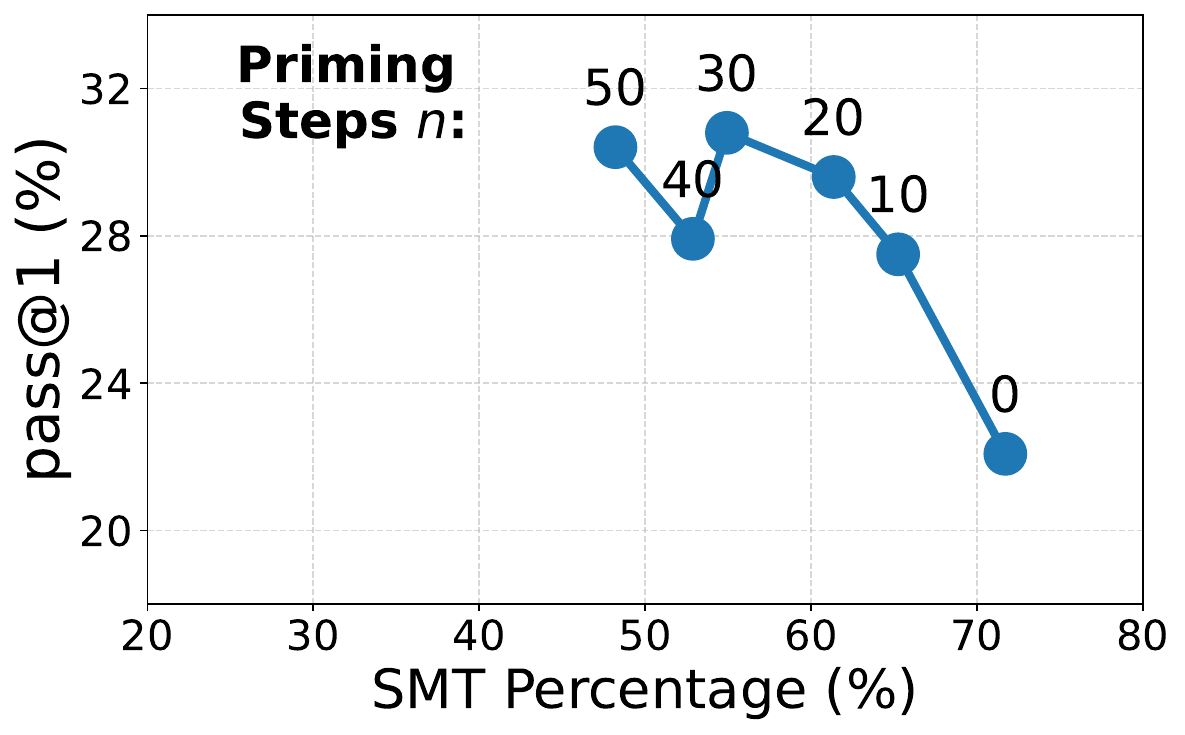}
        \caption{}
        \label{fig:ab2}
    \end{subfigure}
    \hfill
    \begin{subfigure}[t]{0.32\textwidth}
        \centering
        \includegraphics[scale=0.235]{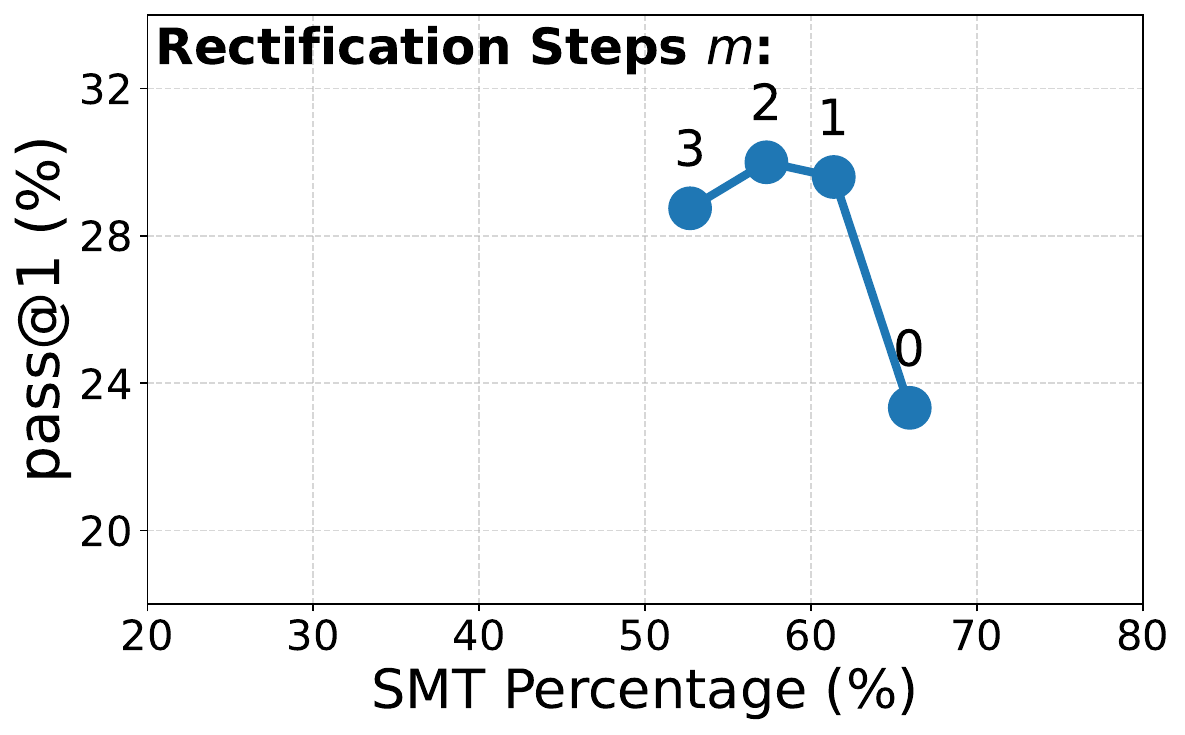}
        \caption{}
        \label{fig:ab3}
    \end{subfigure}
    \vspace{-8pt}
    \caption{
Ablation studies on TrigReason's three triggers and their hyperparameters.
}
\vspace{-15pt}
    \label{fig:ablation}
\end{figure*}


\subsection{Evaluation in Edge-Cloud Collaboration}

To assess TrigReason in realistic deployment scenarios, we simulate an edge-cloud setup: the SRM (DeepSeek-R1-1.5B) runs locally, while the LRM is accessed remotely via the DeepSeek API~\citep{deepseek_api}, which internally uses DeepSeek-V3.1. Latency and API cost are reported in Figure~\ref{fig:overhead1} and Figure~\ref{fig:overhead2}; here, we present accuracy results on AIME24 in Figure~\ref{fig:edge_co}.

TrigReason successfully offloads up to 59.4\% of reasoning tokens to the SRM, while requiring the LRM to generate only 40.6\% steps, with just a 2.49\% absolute accuracy drop compared to full LRM execution. In contrast, SpecReason suffers from degraded accuracy despite high SRM token usage, due to unreliable verification by the LRM, as analyzed in Section~\ref{sec:unreliability}. We observe that when acting as verifier, DeepSeek-V3.1 often assigns high scores to semantically weak or incomplete SRM-generated steps, likely influenced by its own inductive biases in reasoning style. This leads to acceptance of invalid speculative steps, enabling error propagation and ultimately compromising solution correctness.

\subsection{Ablation Study}
\label{sec:ablation}

To evaluate the contribution of each component in TrigReason, we conduct ablation studies on the three proposed triggers and their associated hyperparameters. We also provide a more detailed statistical analysis of their trigger activation frequencies in Appendix~\ref{app:trigger_stats}.

\vspace{-5pt}
\paragraph{Cognitive Overload Threshold $\rho$.}
We analyze $\rho$, which governs activation of the \textit{Cognitive Offload Trigger}. Lower $\rho$ prompts earlier LRM intervention; $\rho = 1$ disables the trigger entirely. We fix $n = 20$, $m = 1$, and use two representative model pairs: DeepSeek-R1-1.5B + QwQ-32B and Qwen3-0.6B + Qwen3-30B-A3B-Thinking.

As shown in Figure~\ref{fig:ab1}, disabling the Cognitive Offload Trigger ($\rho = 1$) causes a significant accuracy drop, confirming its critical role in preventing error accumulation when the SRM exceeds its capacity. Crucially, the optimal $\rho$ is model-dependent: DeepSeek-R1-1.5B + QwQ-32B pair achieves peak performance at $\rho = 0.85$, while Qwen3-0.6B + Qwen3-30B-A3B-Thinking pair performs more stably at  $\rho = 0.75$. This reflects intrinsic SRM differences in average perplexity and reasoning reliability. While higher $\rho$ improves accuracy, it reduces SMT percentage, trading off efficiency. Thus, $\rho$ acts as a tunable knob balancing accuracy and efficiency based on the specific SRM-LRM pair.

\vspace{-5pt}
\paragraph{Priming Steps $n$.}
We vary $n$ from 0 to 50 (with$\rho = 0.85$ and $m = 1$) to assess the \textit{Strategic Priming Trigger}, which enables LRM-provided planning before SRM execution.

Figure~\ref{fig:ab2} shows that reducing $n$ from 20 to 0 incurs a 25.4\% absolute accuracy drop, underscoring the importance of strategic guidance in enabling autonomous SRM reasoning. However, increasing $n$ beyond 30 yields diminishing returns and degrades efficiency. This indicates that early strategic guidance is critical, while excessive priming wastes LRM capacity on execution.

\vspace{-5pt}
\paragraph{Rectification Steps $m$.}
We evaluate the \textit{Intervention Request Trigger} by varying $m$, the number of LRM-generated steps after detecting stagnant reasoning loops. We fix $\rho = 0.85$, $n = 20$.

Figure~\ref{fig:ab3} shows that $m = 1$ already recovers most of the performance gap, with marginal gains from $m = 2$ or $m = 3$. This suggests that the LRM’s corrective capability is highly concentrated: a single high-quality step often suffices to realign the reasoning path. Larger $m$ unnecessarily increases LRM usage and reduces efficiency, making $m = 1$ the optimal trade-off in practice.

\subsection{Performance on Diverse Reasoning Domains}
\label{app:generalization}

To evaluate the generalizability of TrigReason beyond mathematical reasoning, we conduct experiments on two additional benchmarks: Big-Bench Hard (BBH)~\cite{suzgun-etal-2023-challenging}, which focuses on complex logical reasoning, and the AI2 Reasoning Challenge (ARC)~\cite{allenai:arc}, which requires commonsense and scientific knowledge. We use the same model pair as in the main experiments—DeepSeek-R1-1.5B as the SRM and QwQ-32B as the LRM. Since BBH and ARC are less computationally demanding than AIME24, we reduce the number of priming step in strategic priming trigger from 20 to 5 for efficiency. All other hyperparameters, including the cognitive overload threshold ($\rho = 0.85$, corresponding to perplexity $< 1.05$) and the rectification steps intervention request mechanism ($m = 1$), are kept unchanged from the original math reasoning setup, without any domain-specific tuning.

The results are shown in Table~\ref{tab:cross_domain}. TrigReason achieves an accuracy of 0.687 on BBH and 0.948 on ARC-Challenge, outperforming both the SRM alone and SpecReason. Notably, it even slightly exceeds the performance of LRM inference while using the LRM only sparingly. The consistent gains across math, logical, and commonsense reasoning suggest that our heuristics capture general signals of reasoning difficulty rather than overfitting to a specific domain. This supports the robustness and transferability of TrigReason’s design.

\begin{table}[h]
\centering
\caption{Performance comparison on BBH and ARC dataset. All methods use DeepSeek-R1-1.5B as SRM and QwQ-32B as LRM.}
\resizebox{1\columnwidth}{!}{
\label{tab:cross_domain}
\begin{tabular}{lcccc}
\toprule
Dataset & SRM & LRM & SpecReason & TrigReason \\
\midrule
BBH & 0.422 & 0.675 & 0.663 & 0.687 \\
ARC & 0.593 & 0.957 & 0.942 & 0.948 \\
\bottomrule
\end{tabular}}
\end{table}

\subsection{Performance under Varying Token Budgets}
\label{sec:token_budget}

\begin{figure}[!htb]
\centering
\includegraphics[scale=0.35]{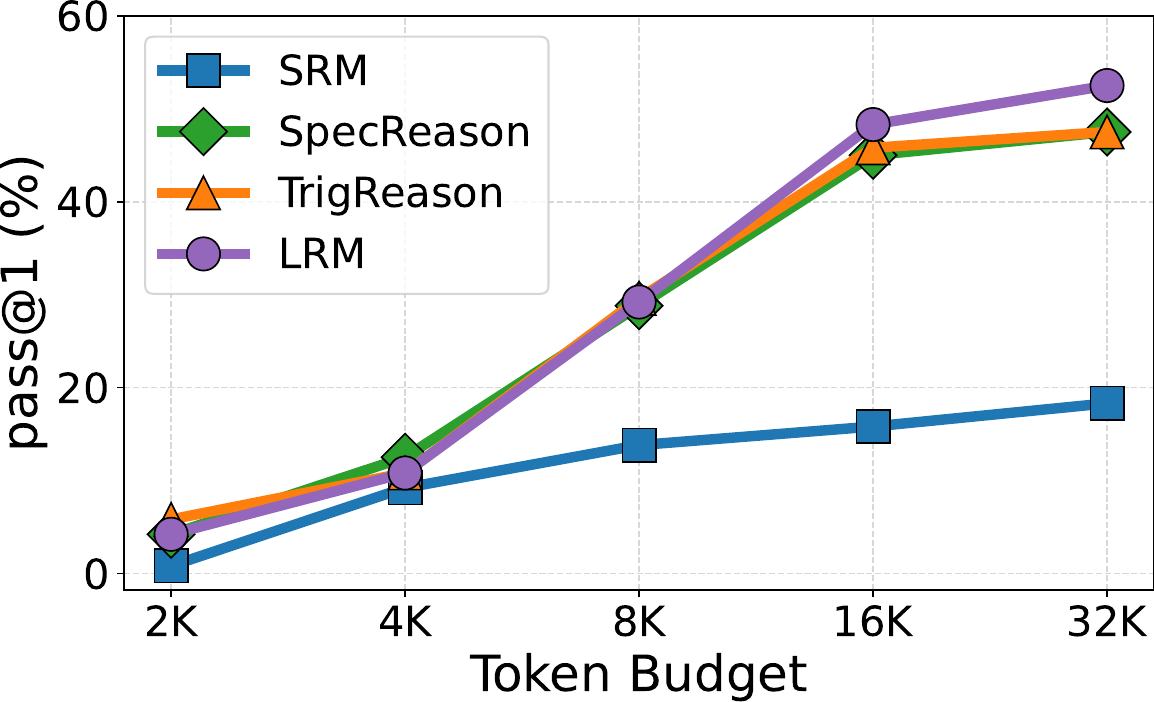}
\caption{Accuracy comparison under different token budgets.}
\label{fig:budget}
\end{figure}

We evaluate the effect of varying thinking token budgets (2K, 4K, 8K, 16K, 32K) on performance using AIME24. As shown in Figure~\ref{fig:budget}, TrigReason consistently outperforms the SRM-only baseline across all settings and matches the performance of both LRM-only and SpecReason, demonstrating its effectiveness and generalization under constrained reasoning resources.

However, as the budget increases, TrigReason and SpecReason exhibit a relative performance gap compared to LRM-only reasoning. This suggests that while collaborative reasoning is highly efficient at lower budgets, it may introduce slight suboptimality when targeting very peak accuracy in resource-abundant settings.


\section{Conclusion}

We propose TrigReason, a collaborative reasoning framework between small and large reasoning models, which achieves better trade-off among accuracy, efficiency, and cost through a trigger-based mechanism that introduces sparse yet critical LRM interventions.




\section*{Limitations}

Although TrigReason provides a practical and generalizable approach to improving the efficiency of reasoning models, certain aspects of its trigger design rely on heuristic criteria. For instance, the cognitive offload trigger is based on model overconfidence. However, the relationship between overconfidence and actual reasoning errors remains insufficiently understood and warrants further investigation. Additionally, like most speculative inference methods, TrigReason primarily targets latency reduction. This comes at the memory overhead from the small reasoning model, which may limit its applicability in memory-constrained deployment environments.

\bibliography{custom}

\clearpage
\appendix
\section{Detailed analysis of LRM Judgment}
\label{app:judge}

To further investigate the unreliability of the LRM-as-a-Judge mechanism in SpecReason, we conduct a fine-grained analysis of rejection behavior on the AIME24 benchmark. Specifically, we compare the step-level rejection rates of two conditions across all 30 questions: (1) the original SpecReason setup, where the LRM judges SRM-generated reasoning steps, and (2) an LRM-own control, where the LRM judges its own reasoning trajectory generated during full LRM execution.

\begin{figure*}[!h]
\centering
\includegraphics[width=0.9\linewidth]{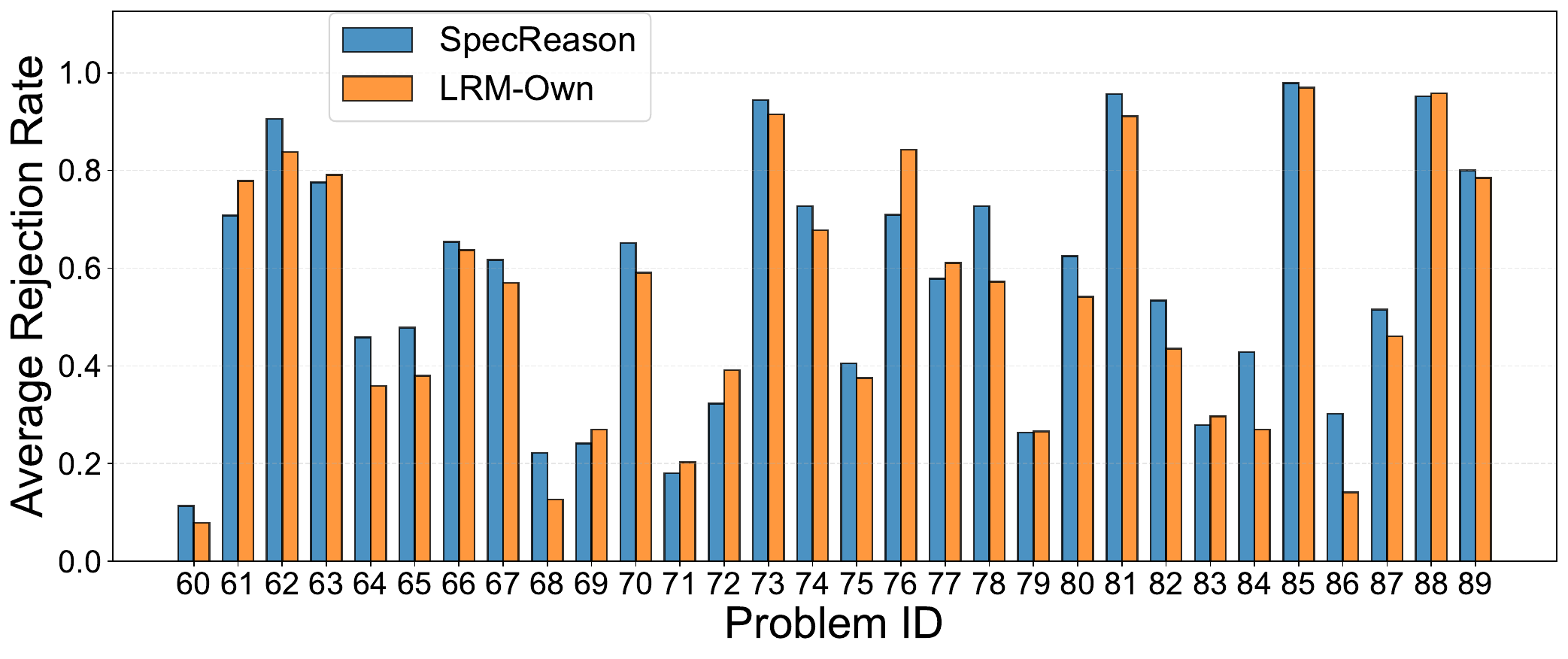}
\caption{Per-question rejection rates of SRM-generated steps (SpecReason) vs. LRM-generated steps (LRM-own) on AIME24. The LRM frequently rejects its own valid reasoning, revealing the inconsistency of its judgment.}
\label{fig:app_reject_rate}
\end{figure*}

Figure~\ref{fig:app_reject_rate} presents the per-question rejection rates for both settings. Despite the semantic correctness of its own reasoning path, the LRM rejects its own steps at a rate comparable to that of SRM-generated steps—indicating inconsistent and preference-driven judgment. On average, the LRM rejects 53.4\% of its own reasoning steps, only slightly lower than the 56.8\% rejection rate for SRM steps. This high self-rejection rate suggests that the LRM's scoring mechanism is not grounded in logical validity, but rather in stylistic or strategic preferences.

This finding strongly supports our claim in Section~\ref{sec:unreliability}: using the LRM as a judge introduces inherent unreliability, as it cannot reliably distinguish between valid reasoning variations and genuinely erroneous steps. Consequently, SpecReason may reject correct SRM reasoning or accept flawed ones based on superficial alignment, undermining both efficiency and correctness.

\section{Case Studies of Three Risk Patterns}
\label{app:case_risk}

To provide intuitive illustrations of the three key risk patterns identified in Section~\ref{sec:risky}, we present visual case studies on representative problems from the AIME24 benchmark.

\begin{figure*}[ht]
\centering
\includegraphics[width=0.85\linewidth]{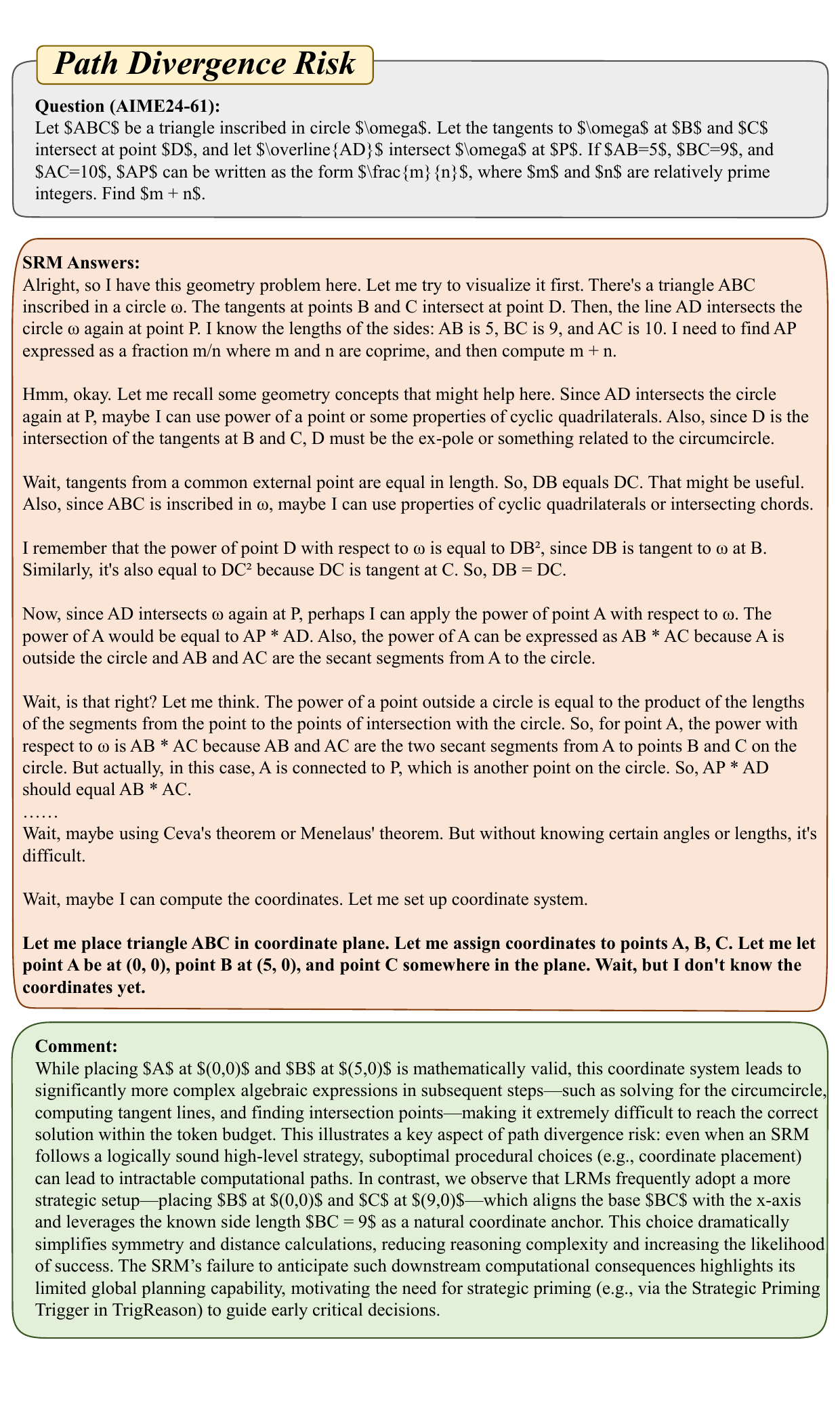}
\caption{Case study of Path Divergence Risk.}
\label{fig:case_study_1}
\end{figure*}

\begin{figure*}[ht]
\centering
\includegraphics[width=0.85\linewidth]{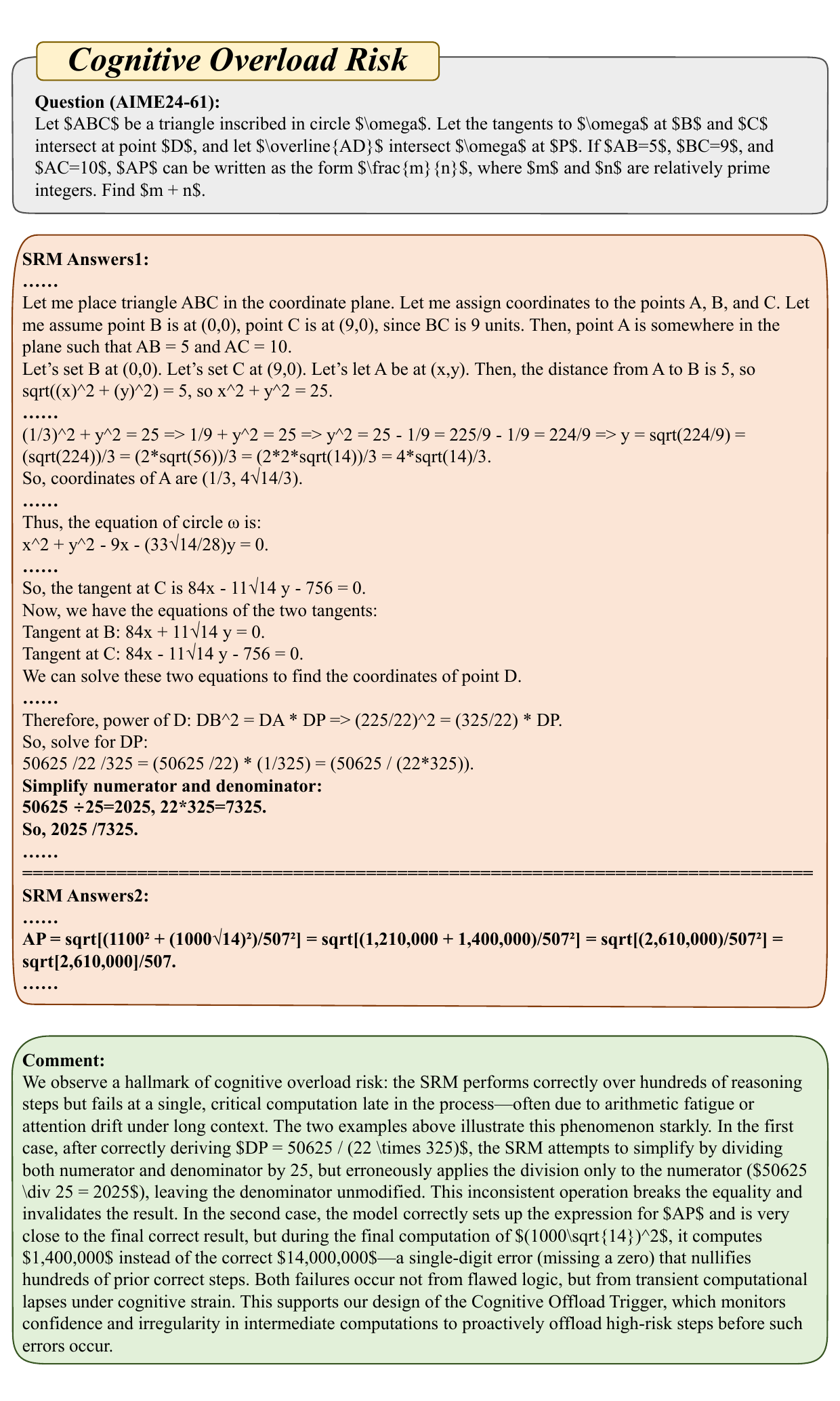}
\caption{Case study of Cognitive Overload Risk.}
\label{fig:case_study_2}
\end{figure*}

\begin{figure*}[ht]
\centering
\includegraphics[width=0.85\linewidth]{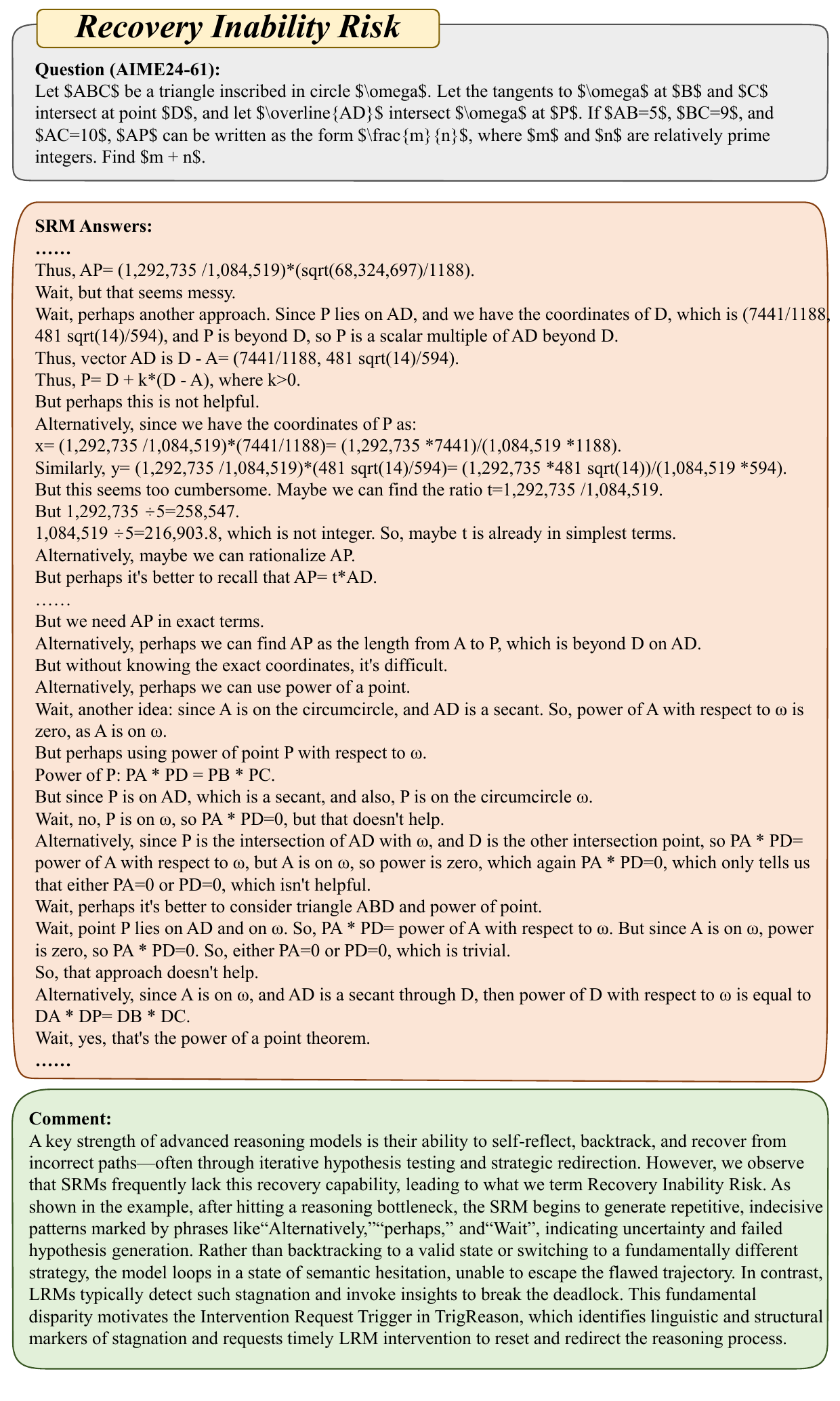}
\caption{Case study of Recovery Inability Risk.}
\label{fig:case_study_3}
\end{figure*}

\begin{itemize}
\item Figure~\ref{fig:case_study_1} demonstrates the Path Divergence Risk, where the SRM makes suboptimal procedural choices that lead to intractable computation, while the LRM adopts a more strategic formulation.
\item Figure~\ref{fig:case_study_2} illustrates Cognitive Overload Risk: the SRM performs correctly for hundreds of steps but fails at a critical late-stage computation due to arithmetic or attentional lapse.
\item Figure~\ref{fig:case_study_3} showcases Recovery Inability Risk, where the SRM enters a loop of indecisive reasoning after hitting a bottleneck, failing to self-correct or switch strategies.
\end{itemize}

\section{Statistics of Trigger Activation}
\label{app:trigger_stats}

To better understand the efficiency and contribution of the three trigger mechanisms in TrigReason, we conduct a quantitative analysis of their activation frequencies and correlation with final task accuracy on the AIME24 dataset.

Table~\ref{tab:trigger_activation} reports the percentage of reasoning steps that activate each trigger under various configurations of cognitive overload threshold $\rho$, priming steps $n$, and rectification steps $m$. We also include the corresponding final accuracy on AIME24 for each setting.

\begin{table*}[h]
\caption{Activation frequencies of triggers and corresponding accuracy on AIME24 dataset. All percentages are computed over total reasoning steps.}
\resizebox{2.1\columnwidth}{!}{
\centering
\label{tab:trigger_activation}
\begin{tabular}{lccccc}
\toprule
Config ($\rho$-$n$-$m$) & Cognitive Offload (\%) & Strategic Priming (\%) & Intervention Request (\%) & Total Trigger (\%) & AIME24 ACC \\
\midrule
0.75-20-1 & 38.50 & 8.06 & 4.36 & 50.92 & 30.0 \\
0.85-20-1 & 25.95 & 8.31 & 4.73 & 38.99 & 29.6 \\
0.95-20-1 & 11.49 & 8.64 & 4.95 & 25.08 & 26.25 \\
0.85-10-1 & 26.41 & 4.23 & 4.87 & 35.51 & 27.5 \\
0.85-20-0 & 26.04 & 8.23 & 0.00 & 34.27 & 23.33 \\
\bottomrule
\end{tabular}}
\end{table*}

We find that the cognitive offload trigger is the most frequently activated mechanism, accounting for over 25\% of all steps in most configurations. The Intervention Request Trigger has a lower activation rate but contributes significantly to performance gains, enabling a performance jump from 23.33 to 29.6 accuracy when enabled (0.85-20-0 vs. 0.85-20-1), despite being triggered in only ~4.7\% of steps. Increasing trigger frequency generally improves accuracy, but with diminishing returns when it's close to LRM performance. For example, raising the cognitive offload threshold from 0.75 to 0.85 reduces its activation by ~12\%, yet accuracy remains nearly unchanged.

\section{Overconfidence as a Sign of Cognitive Overload}
\label{app:overconfidence}

We analyze 160 reasoning trajectories from SRMs across 10 problems where SRMs and LRMs exhibit significant performance gaps. In this analysis, we identify 93 trajectories containing clearly incorrect reasoning steps. A striking pattern emerges: these erroneous steps frequently exhibit overconfidence. To quantify this, we compute the proportion of \emph{low-perplexity tokens} (defined as tokens with per-token perplexity $< 1.05$) within each reasoning step. Among the 93 trajectories with identifiable errors, 88 (94.6\%) contain steps where over 85\% of tokens are low-perplexity. In contrast, only 38.1\% of all reasoning steps in the full set exceed this threshold.

This stark discrepancy suggests that high confidence in SRM outputs is not indicative of correct or deep reasoning, but rather reflects a tendency to fall back on memorized patterns from training data. We present representative examples in Table~\ref{tab:overconfidence_examples2} and Table~\ref{tab:overconfidence_examples1}, where seemingly confident steps lead to incorrect conclusions despite low token-level perplexity.

We interpret this overconfidence as a symptom of \textbf{cognitive overload}: when faced with challenging reasoning junctures, the SRM fails to engage in exploratory or reflective thinking and instead generates superficially fluent but semantically shallow continuations, effectively giving up by defaulting to familiar sequences. This behavior motivates our design of the Cognitive Offload Trigger in TrigReason, which detects such states and delegates to a more capable model before critical errors occur.

\section{Complete List of hesitation words}
\label{app:hesitation}

To operationalize the linguistic markers of Recovery Inability Risk, we define a set of hesitation words and phrases that indicate uncertainty, self-doubt, or backtracking in reasoning trajectories. These patterns are used to detect when the SRM enters a state of semantic hesitation. 

The hesitation word list is derived through a two-stage analysis of 960 reasoning traces generated by DeepSeek-R1-1.5B and Qwen3-0.6B on the AIME24 dataset. First, we identified reasoning steps exhibiting hesitation or stagnation by prompting LRM like QwQ-32B. And then we performed n-gram frequency analysis and compiled a candidate set of high-frequency patterns. Finally, we manually curated this candidate set to retain only those expressions that consistently convey hesitation in context. We implement a case-insensitive regular expression matcher to identify such expressions in generated text. The full list of hesitation patterns is shown in Table~\ref{tab:hesitation_words}.

\begin{table*}[!htb]
\centering
\caption{List of hesitation words and phrases.}
\label{tab:hesitation_words}
\small
\begin{tabular}{@{}>{\ttfamily}l @{\quad} >{\ttfamily}l @{\quad} >{\ttfamily}l@{}}
\toprule
\textnormal{\textbf{Word/Phrase}} &  &  \\
\midrule
wait        & hmm           &   debatable     \\
maybe       & perhaps       & could be          \\
might be    & possibly      & on the other hand \\
alternatively & another possibility & or perhaps     \\
actually    & now that I think about it & I think I made a mistake \\
let me reconsider & not sure & I'm not entirely sure \\
this might be wrong & I could be mistaken & unless I'm wrong \\
\bottomrule
\end{tabular}
\end{table*}

\section{The Main Algorithm of TrigReason}
\label{sec:algorithm}

The main algorithm of TrigReason is shown in Algorithm \ref{alg:trigger_reasoning}.

\renewcommand{\algorithmicrequire}{\textbf{Input:}}
\renewcommand{\algorithmicensure}{\textbf{Output:}}

\begin{algorithm*}
\caption{TrigReason}
\label{alg:trigger_reasoning}
\begin{algorithmic}[1]
\Require 
    Question $x$, small reasoning model $S$, large reasoning model $L$, priming steps $n$, overload threshold $\rho$, rectification steps $m$
\State Initialize: $y \gets [\,]$, $\text{rectify\_step} \gets 0$, $t \gets 0$
\While{not \text{finished}}
    \State $t \gets t + 1$
    
    \If{$t < n$} \Comment{Strategic Priming Trigger}
        \State $y_t \sim p_L(\cdot \mid y_{<t}, x)$
    \Else
        \State $\left(y_t^S, \text{finished}, \text{ppl\_ratio}_t\right) \gets \text{GenerateStep}(S, x, y_{<t})$
        \If{$\text{rectify\_step} > 0$ \textbf{or} $\text{ppl\_ratio}_t > \rho$}
            \Comment{Cognitive Offload or Recovery Trigger}
            \State $y_t \sim p_L(\cdot \mid y_{<t}, x)$
            \If{$\text{ppl\_ratio}_t <= \rho$}
                \State $\text{rectify\_step} \gets \text{rectify\_step} - 1$
            \EndIf
        \Else
            \State $y_t \gets y_t^S$ \Comment{Accept small model output}
            \If{$\text{Detect\_hesitation}(y_t, y_{t-1}, y_{t-2}))$}
                \State $\text{rectify\_step} \gets m$
                    \Comment{Intervention Request Trigger fires}
            \EndIf
        \EndIf
    \EndIf
    
    \State Append $y_t$ to $y$
\EndWhile

\State \Return $y$
\Ensure 
    Reasoning trajectory $y$, final answer
\end{algorithmic}
\end{algorithm*}

\begin{table*}[!htb]
\centering
\caption{Example of incorrect reasoning steps with corresponding perplexity ratios (ppl\_ratio).}
\label{tab:overconfidence_examples2}
\begin{tabularx}{\textwidth}{@{}>{\raggedright\arraybackslash}m{0.8\textwidth} c@{}}
\toprule
\textbf{Reasoning Step} & \textbf{ppl\_ratio} \\
\midrule

 AP = sqrt[(1100² + (1000$ \surd $ 14)²)/507²] = sqrt[(1,210,000 + 1,400,000)/507²] = sqrt[(2,610,000)/507²] = sqrt[2,610,000]/507.
 & 0.955 \\ \hline
 \[
\frac{b^2}{1 + m^2} \cdot \frac{1}{120} = 1
\]

So:
\[
b^2 = 120 (1 + m^2)
\]

Therefore, \(b^2 = 120 (1 + m^2)\) and \(a^2 = \frac{120 (1 + m^2)}{6 - 5m^2}\).
& 0.916 \\ \hline
Simplify numerator and denominator:

50625÷25=2025, 22*325=7325.

So, 2025/7325.
& 0.931 \\ \hline
Okay, so from n=1 to n=20, the losing positions (L) are: 

2, 5, 6, 10, 11, 15, 16, 20.
& 0.872 \\ \hline
Let me denote the diagonals as vectors \(\vec{d_1}\) and \(\vec{d_2}\), which are perpendicular. So, if the rhombus is centered at the origin, then the vertices can be expressed as \(\frac{\vec{d_1}}{2}\), \(\frac{\vec{d_2}}{2}\), \(-\frac{\vec{d_1}}{2}\), and \(-\frac{\vec{d_2}}{2}\).
& 0.863 \\ \hline
Finally, when her walking speed is \( s + \frac{1}{2} = 3 \) km/h, the time taken is:
\[
\frac{9}{3} + t = 3 + t
\]
Since \( t = 4 - 3 = 1 \) hour, the total time is:
\[
3 + 1 = 4 \text{ hours}
\]
& 0.962 \\ \hline
For the second scenario:
- Walking speed: \( s + 2 \) km/h
- Total time: 2 hours and 24 minutes = 144 + 24 = 168 minutes
- Time in the coffee shop: \( t \) minutes
- Walking time: \( 168 - t \) minutes
& 0.960 \\

\bottomrule
\end{tabularx}
\end{table*}

\begin{table*}[!htb]
\centering
\caption{Example of incorrect reasoning steps with corresponding perplexity ratios (ppl\_ratio).}
\label{tab:overconfidence_examples1}
\begin{tabularx}{\textwidth}{@{}>{\raggedright\arraybackslash}m{0.8\textwidth} c@{}}
\toprule
\textbf{Reasoning Step} & \textbf{ppl\_ratio} \\
\midrule
Subtracting Equation 1 from Equation 2:
\[
\frac{540}{s + 2} - \frac{540}{s} = 144 - 240 = -96
\]
\[
540 \left( \frac{1}{s + 2} - \frac{1}{s} \right) = -96
\]
\[
540 \left( \frac{-2}{s(s + 2)} \right) = -96
\]
\[
\frac{1080}{s(s + 2)} = -96
\]
 & 0.974 \\ \hline
 \[
 V = (sqrt( (a² + b² - c²)(a² + c² - b²)(b² + c² - a²) )) / (6 * sqrt(2))
 \]
& 0.952 \\ \hline

I can eliminate \( t \) and find the value of \( s \):
\[
\frac{9}{s} - \frac{9}{s + 2} = 1.6
\]
Simplifying this, I get:
\[
\frac{18}{s(s + 2)} = 1.6
\]
Solving for \( s \), I find:
\[
s = 5 \text{ km/h}
\]
 & 0.876 \\ \hline
Finally, when walking at \( s + \frac{1}{2} = 3 \) km/h, the time taken for the walk is:
\[
\frac{9}{3} = 3 \text{ hours} = 180 \text{ minutes}
\]
 & 0.935 \\ \hline
 n | Outcome (Winning or Losing for Alice)
---|---
1 | Winning (Alice takes 1)
2 | Losing (Alice leaves Bob with 1)
3 | Winning (Alice takes 1, leaving 2 for Bob)
4 | Winning (Alice takes 4)
5 | Losing (Alice leaves Bob with 4 or 1)
6 | Losing (Alice leaves Bob with 5 or 2)
7 | Winning (Alice leaves Bob with 6 or 3)
8 | Winning (Alice leaves Bob with 7 or 4)
9 | Losing (Alice leaves Bob with 8 or 5)
10 | Losing (Alice leaves Bob with 9 or 6)
11 | Winning (Alice leaves Bob with 10 or 7)
12 | Winning (Alice leaves Bob with 11 or 8)
13 | Losing (Alice leaves Bob with 12 or 9)
14 | Losing (Alice leaves Bob with 13 or 10)
15 | Winning (Alice leaves Bob with 14 or 11)...

 & 0.929 \\

\bottomrule
\end{tabularx}
\end{table*}

\section{The Use of Large Language Models}
\label{sec:appendix_llm_use}

In this work, large language models are used exclusively to assist with language editing and clarification during the writing of this paper. All technical ideas, method design, analysis, and experimental work are conducted by human authors.

\end{document}